%% file: main_preprint.tex
\PassOptionsToPackage{table}{xcolor}
\PassOptionsToPackage{protrusion=true,expansion=true}{microtype}
\documentclass[]{aidata}

\usepackage{amsmath,amssymb}
\usepackage{booktabs}
\usepackage{multirow}
\usepackage{colortbl}
\usepackage{float}
\usepackage{url}
\usepackage{pifont}
\titleformat*{\paragraph}{\bfseries}
\usepackage{textcomp}
\tcbuselibrary{listings}
\newtcblisting{promptbox}[2][breakable]{enhanced, #1, listing only,
  colback=black!2, colframe=black, colbacktitle=black, coltitle=white,
  fonttitle=\sffamily\bfseries\small, title={#2},
  boxrule=0.6pt, arc=1mm, left=2mm, right=2mm, top=1mm, bottom=1mm,
  before skip=6pt, after skip=8pt,
  listing options={basicstyle=\ttfamily\scriptsize, breaklines=true, columns=fullflexible, keepspaces=true, showstringspaces=false, upquote=true, aboveskip=0pt, belowskip=0pt}}

\DeclareRobustCommand{\arxivtight}{\looseness=-1\relax}
\DeclareRobustCommand{\arxivfill}{{\parfillskip=0pt plus 0.48\hsize\tolerance=1000\emergencystretch=1em\relax\par}}
\DeclareRobustCommand{\arxivfilltight}{{\looseness=-1\parfillskip=0pt plus 0.48\hsize\tolerance=1000\emergencystretch=1em\relax\par}}

\title{CoEvoBench: Can LLMs Drive Recursive Co-Evolution of Agentic Systems?}

\author[1,2,\ast]{Chuan Xiao}
\author[1,2,\ast]{Baojie Wang}
\author[2,3]{Haizhao Fan}
\author[2,\dagger]{Hongzhe Du}
\author[2,\dagger]{Xander Xu}

\affiliation[1]{Zhejiang University}
\affiliation[2]{Alibaba Group}
\affiliation[3]{Shanghai Jiao Tong University}

\contribution[\ast]{Equal contribution}
\contribution[\dagger]{Corresponding authors}

\abstract{\input{sections/abstract}}

\correspondence{\email{duhongzhe.dhz@alibaba-inc.com}, \email{yanquan.xx@alibaba-inc.com}}
\checkdata[Code]{\href{https://github.com/AlibabaResearch/CoEvoBench}{\texttt{https://github.com/AlibabaResearch/CoEvoBench}}}

\begin{document}

\maketitle

\input{sections/introduction}
\input{sections/related_work}
\input{sections/benchmark}
\input{sections/experiments}
\input{sections/analysis}
\input{sections/conclusion}

\clearpage

\subsection*{Reproducibility statement}
The full experimental code is released at \url{https://github.com/AlibabaResearch/CoEvoBench}. Appendices~\ref{app:benchmarks}--\ref{app:integrity} describe the task splits, initial and reference harnesses, research instructions, and evaluation settings, and Appendices~\ref{app:repeat}--\ref{app:audit} report the repeated-evaluation noise and the audit procedure.

\bibliographystyle{unsrtnat}
\bibliography{references}

\clearpage

\beginappendix

\input{sections/appendix_related}

\input{sections/appendix}

\input{sections/appendix_component_comparisons}

\input{sections/appendix_evaluation}

\end{document}

%% file: sections/abstract.tex
The performance of an LLM agent depends on both its model and its harness, the code that mediates its interaction with tools and environments.
Can LLM researchers drive this co-evolution as a recursive improvement process, deciding round after round which component to change when each change alters the value of the other?
Existing benchmarks evaluate post-training or data research with the harness out of scope, or harness optimization with the model fixed, leaving this joint decision unmeasured.
We introduce CoEvoBench, a benchmark that evaluates LLMs as researchers co-evolving the model and harness of a target agent across five agentic domains.
Researchers start from a common base model and initial harness, train and revise within a 24-hour budget, and submit a model--harness pair that is scored on separate test tasks.
Across five frontier researchers, the strongest closes more than four fifths of the gap to an instruction-tuned reference system, and co-evolution succeeds on interaction tasks but rarely on long-horizon coding.
These results suggest that current LLMs can make a strong first improvement, but cannot yet turn feedback from an improved system into the next one.
CoEvoBench provides a controlled testbed for measuring the research capabilities that recursive co-evolution requires.\arxivfill

%% file: sections/introduction.tex
\section{Introduction}
\label{sec:introduction}

LLM agents now repair software, operate terminals, and carry out workflows in applications and services~\citep{swebenchverified,tblite,appworld,tau2bench,bfcl}, and their performance depends as much on the harness that mediates their interaction with tools and environments as on the model itself~\citep{metaharness}. Building such an agent therefore means improving both. Developers run the agent, read its execution traces, decide whether a failure calls for a harness fix or more training, and take the revised system as the starting point of the next round. Because each change to one component alters the value of the other and shifts the system's failure modes, the hard part is keeping this loop going, turning feedback from an already improved system into the next gain. The loop is still run largely by hand. An LLM researcher that could run it would automate agent development and realize a concrete, measurable form of recursive self-improvement (RSI)~\citep{dgm,rsibenchdata,metarsi}. This raises a central question: \textbf{Can LLM researchers drive this recursive improvement process and achieve further system gains through successive iterations?}\arxivfill

\begin{figure}[t]
    \centering
    \includegraphics[width=\linewidth]{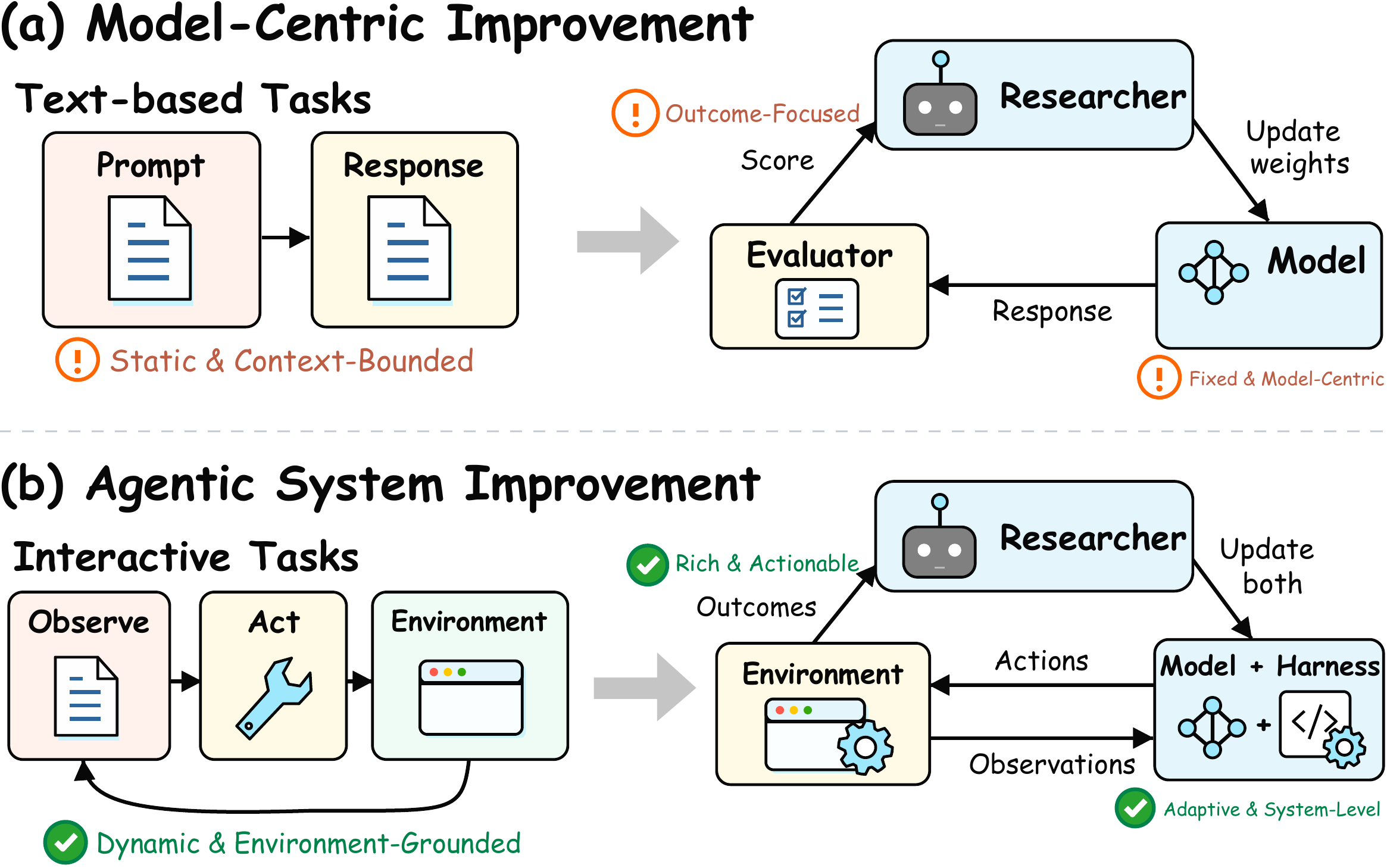}
    \caption{\textbf{Motivation for CoEvoBench.} \textbf{(a) Model-centric improvement:} for reasoning tasks, the researcher uses task feedback to update model weights and improve reasoning. \textbf{(b) Agentic-system-centric improvement:} for agentic tasks, CoEvoBench evaluates researchers that use feedback from task execution and environment interaction to improve model weights, the harness, or both. The harness governs how the model interacts with tools and the environment.\arxivfill}
    \label{fig:motivation}
\end{figure}

Existing benchmarks cannot answer this question, because each covers only one side of the loop. PostTrainBench~\citep{posttrainbench} measures post-training on reasoning tasks, RSIBench-Data~\citep{rsibenchdata} measures data-centric research for post-training, and related benchmarks target algorithm design and agentic RL engineering~\citep{ai4aibench,agent2rlbench}. All of them leave the execution harness out of scope. Meta-Harness~\citep{metaharness}, DarwinX~\citep{darwinx}, and related systems~\citep{dgm,modularrsi} optimize the harness around a fixed model. Recent co-evolution methods update both components through specific pipelines~\citep{sia,coharness,harnessforge,sciencebuddy}, but do not evaluate whether LLM researchers can drive this process on their own. As Figure~\ref{fig:motivation} illustrates, model-centric improvement updates weights alone, while improving an agentic system updates both the model and the harness. None of these works tests the key challenge of recursive improvement. No existing benchmark gives an LLM researcher both levers and asks it to turn feedback from an already improved system into further gains, round after round.

To address this challenge, we introduce \textbf{CoEvoBench}, a benchmark that places LLM researchers in the agent developer's position. A researcher drives the recursive loop end to end, training the model, revising the harness, and choosing each next step from development feedback, and the benchmark records both the system it delivers and the trajectory that produced it. CoEvoBench thus measures how much a researcher improves an agentic system and how its gains arise over successive rounds.\arxivfill

\begin{figure}[t]
    \centering
    \includegraphics[width=\linewidth]{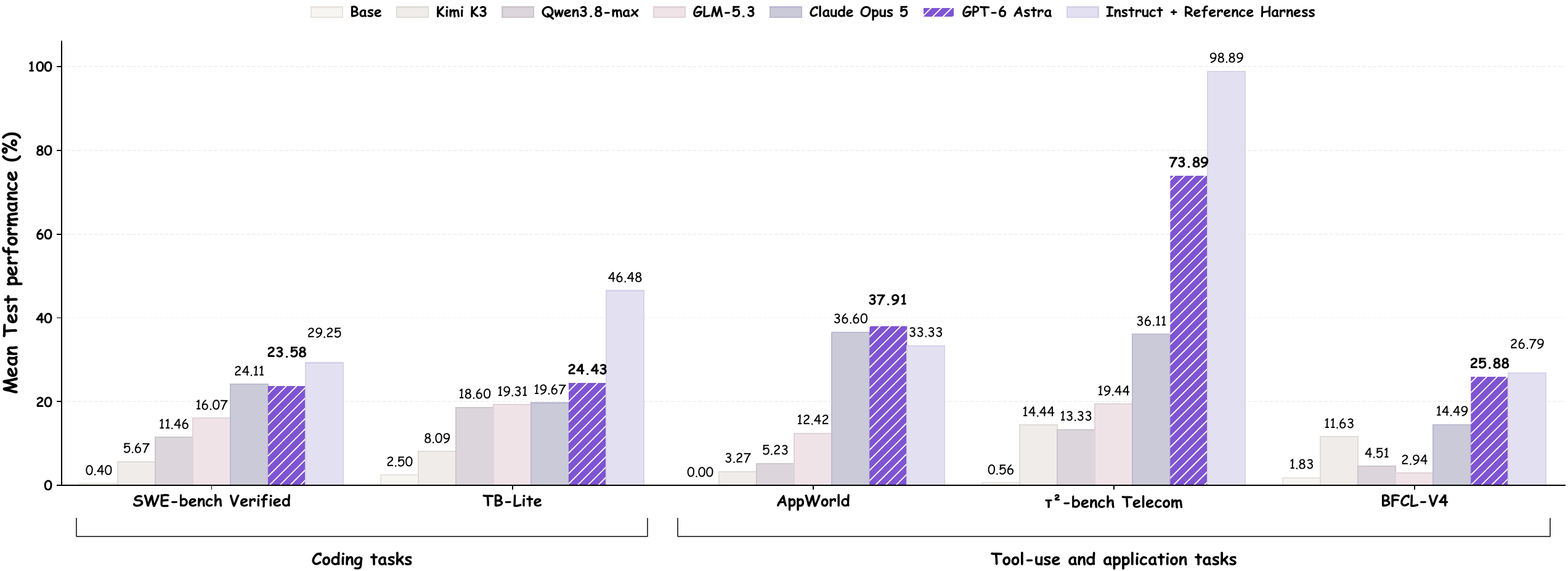}
    \caption{\textbf{CoEvoBench main results.} Mean test performance (\%) of systems developed by each researcher across five agentic benchmarks, averaged over three development runs per researcher--benchmark pair. Base denotes Qwen3.5-9B-Base with the initial harness. Instruct + Reference Harness denotes Qwen3.5-9B-Instruct with a reference harness. Names identify the researcher models.\arxivfill}
    \label{fig:main-results-overview}
\end{figure}

CoEvoBench spans five agentic domains: software repair, terminal operation, application interaction, interactive service workflows, and multi-turn tool use with memory. In each domain, researchers start from a common base model and initial harness, train and revise within a fixed compute budget, and submit a model--harness pair that is scored on separate evaluation tasks. We evaluate five frontier LLMs as researchers, each developing a system three times in every domain on two target models from different model families, Qwen3.5-9B-Base and MiniCPM5-2B-Base. Beyond the delivered systems, we analyze the development trajectories: which component each step changes, how much each candidate adds, and whether results change the next decision. Figure~\ref{fig:main-results-overview} summarizes the mean evaluation-task performance on Qwen3.5-9B-Base across the five domains, alongside the initial Base system and the Instruct model with a reference harness.\arxivfill

Our main contributions are:
\begin{itemize}
    \item We formulate agent development as a measurable instance of recursive self-improvement. An LLM researcher holds both the training and the harness lever and must keep turning feedback from an already improved system into further gains.
    \item We build CoEvoBench across five agentic domains and two target models from different model families. Researchers start from a shared base model and initial harness within a fixed budget, compare candidates using score-only feedback on a selection set, and are scored on separate evaluation tasks (Section~\ref{sec:benchmark-design}).
    \item We show that task characteristics decide whether current LLM researchers can achieve co-evolution, and to what degree. After early breakthroughs, researchers rarely turn feedback from an improved system into further gains, and the loop breaks where results should decide the next step (Section~\ref{sec:analysis}).
\end{itemize}

%% file: sections/related_work.tex
\section{Related Work}
\label{sec:related-work}

\paragraph{Automated AI research and evaluation.} Automated systems conduct scientific research \citep{aiscientist}, while benchmarks assess ML experimentation, engineering, expert research, and paper replication \citep{mlagentbench,mlebench,rebench,paperbench}. Recent suites evaluate post-training, data research, algorithm design, and agentic RL \citep{posttrainbench,rsibenchdata,ai4aibench,agent2rlbench}, alongside agent construction under business and serving constraints \citep{hypertau}. The post-training, data, algorithm, and RL suites improve the target model and leave the harness that executes it unchanged.

\paragraph{Harness-level agent improvement.} With fixed model weights, agents improve through feedback and reusable skills \citep{reflexion,voyager}, prompt and program search \citep{gepa,adas}, and evolving control logic and context management \citep{dgm,metaharness,darwinx,modularrsi}. Benchmarks also assess transfer from retained experience and skill libraries to later tasks \citep{pastbench,continualskillbench}.\arxivfill

\paragraph{Agent training and model--harness co-evolution.} Agent training uses trajectory tuning and executable environments \citep{agenttuning,swegym}, with harnessed RL preserving deployment interactions \citep{lightning}. Co-evolution combines feedback-directed intervention and harness-conditioned training \citep{sia,coharness,harnessforge,harnessx}, scientific-agent learning and operator composition \citep{sciencebuddy,metarsi}, and expert correction for model--harness compatibility \citep{coevolving}. These works build systems that update both components. CoEvoBench evaluates how well LLM researchers drive this co-evolution on their own (Appendix~\ref{app:related-comparison}).\arxivfill

%% file: sections/benchmark.tex
\section{CoEvoBench}
\label{sec:benchmark-design}

\begin{figure}[t]
    \centering
    \includegraphics[width=\linewidth]{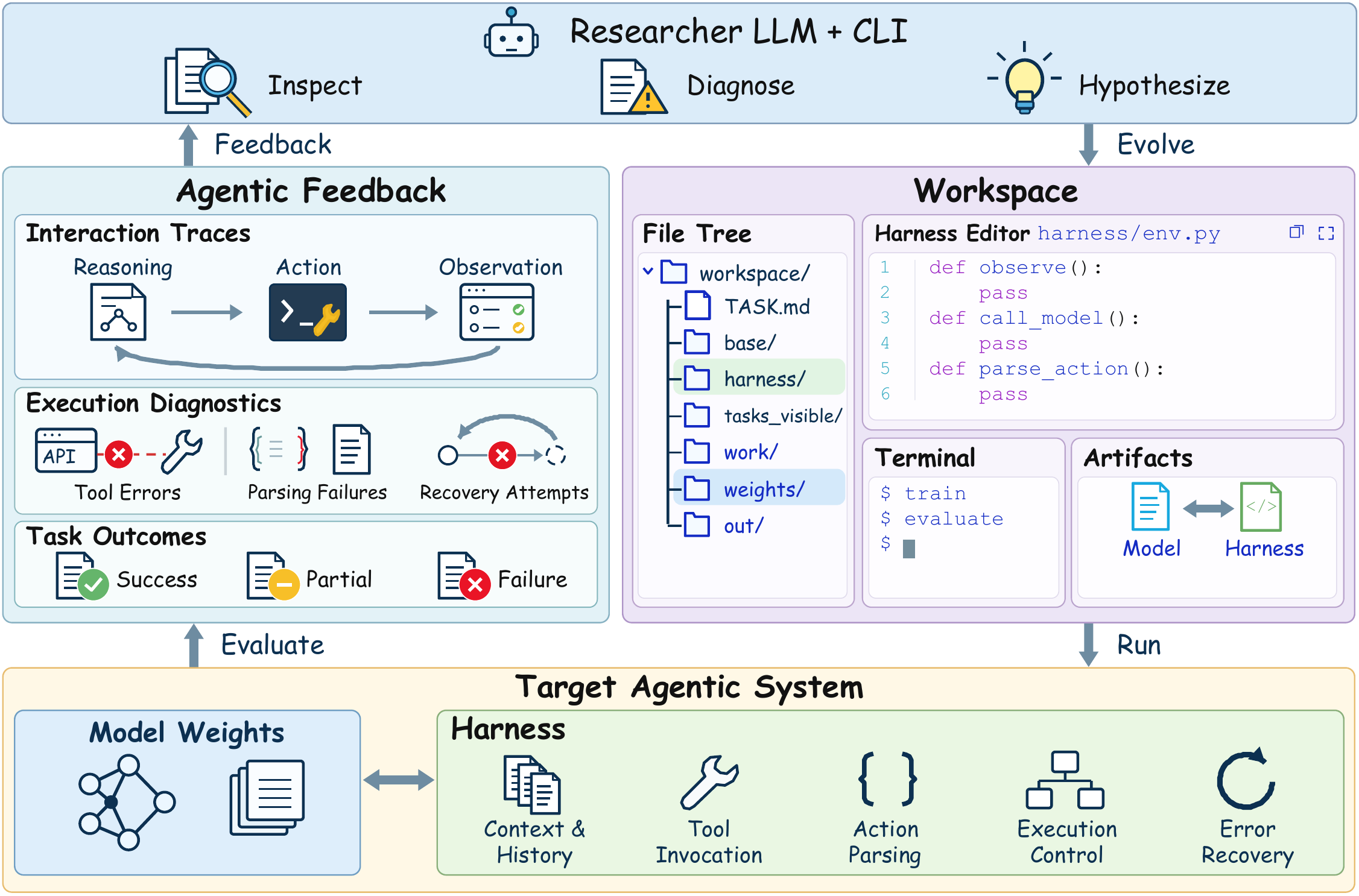}
    \caption{\textbf{An overview of the CoEvoBench development loop.} An LLM researcher improves a target agent consisting of model weights and an executable harness. At each iteration, the researcher uses feedback to diagnose failures, formulate improvement hypotheses, and modify the model weights or the harness. The resulting target agent executes development tasks, producing outcomes and interaction traces. Visible tasks provide task details and execution traces for analysis, while selection-set evaluations return performance feedback without exposing task contents or trajectories. These observations guide the researcher's next modifications, closing the modification--execution--feedback loop. Within a 24-hour development budget, the researcher nominates a final model--harness pair, which is evaluated on separate test tasks without further modification.\arxivfill}
    \label{fig:agentic-system-workflow}
\end{figure}

CoEvoBench evaluates whether a researcher \(R\) can improve a target agent by running the development loop an agent developer would run. The researcher may change any part of the agent, learns only from development feedback, and is judged by the submitted system on unseen tasks. We represent the target agent as \(S=(\theta,H)\), where \(\theta\) denotes model weights and \(H\) is the executable harness governing tool use and environment interaction. For domain \(b\), a development run is specified by
\begin{equation}
\mathcal P_b=\left(S_b^{(0)},\mathcal I_b,\mathcal U_b,\mathcal E_b^{\mathrm{dev}},\mathcal B\right),
\end{equation}
where \(S_b^{(0)}=(\theta_0,H_b^{(0)})\) is the initial system, \(\mathcal I_b\) contains materials accessible to the researcher, \(\mathcal U_b\) is the permitted action space, \(\mathcal E_b^{\mathrm{dev}}\) provides development evaluation feedback, and \(\mathcal B\) is the resource budget. The researcher iterates within this setup and submits a final system \(S_b^\star=(\theta_b^\star,H_b^\star)\), whose improvement is measured through independent testing. Figure~\ref{fig:agentic-system-workflow} connects researcher-driven modifications, target-agent execution, and feedback within the development loop.\arxivfill

\subsection{Benchmark Suite}
\label{sec:domain-instantiations}

The domain set \(\mathcal D\) comprises five agentic benchmarks: SWE-bench Verified~\citep{swebenchverified}, TB-Lite~\citep{tblite}, AppWorld easy~\citep{appworld}, \(\tau^2\)-bench Telecom base subset~\citep{tau2bench}, and BFCL-V4 multi-turn and memory tasks~\citep{bfcl,bfclv4memory}. They cover software repair, terminal operation, application interaction, interactive service workflows, and multi-turn tool use with memory, respectively. We call the first two coding tasks, which need long-horizon edits and command sequences, and the last three interaction tasks, which involve structured exchanges with applications, services, and tools.\arxivfill

For each domain \(b\in\mathcal D\), we partition the task pool \(D_b\) into three disjoint sets:
\begin{equation}
D_b=D_b^v\mathbin{\dot\cup}D_b^s\mathbin{\dot\cup}D_b^t.
\end{equation}
We target a 1:2:2 split, with domain-specific adjustments reflecting task characteristics (Appendix~\ref{app:splits}). The visible set \(D_b^v\) provides material for failure analysis and development. The selection set \(D_b^s\) allows researchers to assess how well their modifications generalize to unseen tasks. The test set \(D_b^t\) supports final evaluation. Accessible materials \(\mathcal I_b\) include initial instructions, system and tool documentation, and visible-set task materials.\arxivfill

We stratify SWE-bench Verified by repository and TB-Lite by difficulty. For AppWorld, we assign each scene's variants to the same split and stratify scenes by their normal or challenge source. Telecom is stratified by operation type. For BFCL, we group multi-turn variants by task family and keep each memory scenario, its cross-backend QA records, and prerequisite conversations in one split.

\subsection{Research Setup and Action Space}
\label{sec:research-setup-action-space}

All runs start from the base target weights \(\theta_0\) (Qwen3.5-9B-Base or MiniCPM5-2B-Base) and a domain-specific initial harness \(H_b^{(0)}\), which provides only a minimal interface to the domain and leaves harness design to the researcher. The resource budget \(\mathcal B\) comprises 24 hours of development, four NVIDIA H20 GPUs, and a \$100 teacher-model calling budget. Researchers allocate GPUs to training and inference.\arxivfill

The action space comprises inspection and evaluation, model updates, and harness modifications:
\begin{equation}
\mathcal U_b=
\mathcal U_b^{\mathrm{inspect/eval}}
\cup\mathcal U_b^{\mathrm{model}}
\cup\mathcal U_b^{\mathrm{harness}}.
\end{equation}
Inspection and evaluation actions cover artifacts, execution feedback, and candidates. Model-update actions construct training data, query the teacher, and update weights \(\theta\). Harness modifications act on \(H\), including tool interfaces, execution control, context management, and memory mechanisms.

Agentic tasks involve multi-step tool and environment interactions, making successful trajectories difficult to obtain from the initial system. We therefore allow limited access to a fixed teacher model, Qwen3.8-27B, for constructing training data through distillation. Teacher assistance is available as a data-construction tool within the model-update action space.\arxivfilltight

\subsection{Development Protocol}
\label{sec:development-evaluation}

During development, researchers can inspect task details and execution trajectories on \(D_b^v\) to diagnose failures, formulate improvement hypotheses, and construct training data. The development-feedback operator \(\mathcal E_b^{\mathrm{dev}}\) supports visible- and selection-set evaluations. Evaluations on \(D_b^s\) return only performance feedback to assess generalization to unseen tasks.\arxivfilltight

At development step \(t\), the researcher selects an action \(a_t\in\mathcal U_b\) based on the interaction history \(\mathcal T_t\), receives an observation \(o_{t+1}\), and incorporates it into subsequent decisions:
\begin{equation}
a_t\sim R(\cdot\mid\mathcal T_t),
\qquad
\mathcal T_{t+1}=\mathcal T_t\oplus(a_t,o_{t+1}).
\end{equation}
Here, \(\mathcal T_0\) contains the initial accessible materials \(\mathcal I_b\), and \(\oplus\) appends each action and its observation to the history. The researcher thus reads feedback, modifies the system, and evaluates candidates. We call this process recursive improvement, since each candidate \(S^{(k+1)}\) is built from the current system \(S^{(k)}\) and the feedback it produces.

The researcher nominates a final system \(S_b^\star=(\theta_b^\star,H_b^\star)\) within the 24-hour development budget and may stop early. Once submitted, the system runs independently on the test set without modification.

CoEvoBench's experiment execution and evaluation infrastructure uses sandbox isolation and benchmark-data access restrictions during development. Researchers modify models and harnesses in isolated workspaces with network access enabled, while datasets that directly expose benchmark task data are blocked. Visible-set materials are supplied through the platform. Selection and test data remain protected, and selection-set task contents and execution trajectories are withheld.\arxivfill

\subsection{Evaluation and Scoring}
\label{sec:evaluation-scoring}
\label{sec:evaluation-integrity}

The platform evaluates the final and initial systems on \(D_b^t\) under the same domain-specific scoring protocol. We measure complete-task success for SWE-bench Verified, mean task reward for TB-Lite, complete-task success for AppWorld and Telecom, and a weighted combination of multi-turn and memory accuracy for BFCL, with weights of 60\% and 40\%, respectively.\arxivfill

After each run, an agent applies a fixed audit skill to the complete development trajectory, covering tool calls, file access, training data lineage, and network logs, and we confirm every violation it flags (Appendix~\ref{app:audit}). For runs with confirmed cheating, we re-evaluate the base model paired with the final submitted harness. Let \(\mathcal E_b^{\mathrm{test}}(S)\) denote the test score of system \(S\) on domain \(b\). The reported score is\arxivtight
\begin{equation}
\widehat{\mathcal E}_b(S_b^\star)=
\begin{cases}
\mathcal E_b^{\mathrm{test}}\!\left((\theta_0,H_b^\star)\right),
& \text{if cheating is confirmed},\\[3pt]
\mathcal E_b^{\mathrm{test}}\!\left((\theta_b^\star,H_b^\star)\right),
& \text{otherwise}.
\end{cases}
\end{equation}
The improvement delivered by the researcher over the initial system is
\begin{equation}
J_b(R)=\widehat{\mathcal E}_b(S_b^\star)-\mathcal E_b^{\mathrm{test}}(S_b^{(0)}).
\end{equation}

We report per-domain test performance and improvements, and summarize cross-domain performance using normalized weights. Let \(\bar s_b(R)\) denote the mean reported test score across three independent development runs of researcher \(R\) on domain \(b\), \(B_b\) the test score of the initial model paired with the initial harness, and \(I_b\) the test score of the instruction-tuned model paired with a reference harness, which anchors each domain's scale through the baseline gap \(I_b-B_b\). We define
\begin{equation}
w_b=\frac{(I_b-B_b)^{-1}}{\sum_{j\in\mathcal D}(I_j-B_j)^{-1}},
\qquad
\mathrm{Avg.}(R)=\sum_{b\in\mathcal D}w_b\bar s_b(R).
\end{equation}
The weights sum to one and are fixed across researchers. Under this weighting, the fraction of the aggregate gap a researcher closes equals its mean fractional gap closure across domains.

%% file: sections/experiments.tex
\section{Experiments}
\label{sec:experiments}

\input{sections/experimental_setup}

\subsection{Main Results}
\label{sec:main-results}

\input{tables/table1_content}

\paragraph{Overall performance.}
LLM researchers substantially improve the target agent, but their effectiveness varies widely (Table~\ref{tab:main-leaderboard}). GPT-6 Astra closes more than four fifths of the gap between the initial system and the Instruct + Reference Harness system, Claude Opus 5 more than half, and the remaining three researchers less than a third. On AppWorld, Astra and Opus both surpass the reference system from an initial score of zero. Every researcher starts from the same system and budget, so this spread reflects differences in research capability. Across the three runs, researchers show a clear gap in capability, with Opus and Astra well ahead of Kimi, Qwen, and GLM. MiniCPM5-2B-Base in Table~\ref{tab:main-leaderboard} shows a similar gradient.\arxivfill

\paragraph{Per-domain performance.}
Where researchers separate depends on the domain. On interaction tasks, the strongest researcher pulls far ahead. Astra closes most of the gap on $\tau^2$-bench Telecom and nearly all of it on BFCL-V4, far more than any other researcher. Coding tasks compress the field. Opus and Astra are tied on SWE-bench Verified, and on TB-Lite four researchers finish within six points of one another, with the best closing only about half the gap. The smaller MiniCPM5-2B-Base target widens this contrast. No researcher exceeds 10\% on SWE-bench Verified or 2\% on TB-Lite, whereas Astra reaches 96.1\% on $\tau^2$-bench Telecom, and Opus and Astra again surpass the reference system on AppWorld. The harness lifts MiniCPM mainly on interaction tasks.\arxivfill

%% file: sections/experimental_setup.tex
\subsection{Experimental Setup}
\label{sec:experimental-setup}

\paragraph{Researcher configurations.}
\label{sec:researcher-configurations}

The five researcher models are GPT-6 Astra, Qwen3.8-max, GLM-5.3, Kimi K3, and Claude Opus 5. Each model runs through a native command-line agent. GPT-6 Astra, with reasoning effort high, runs through Codex, and the other four run through Claude Code. Each model--CLI pair constitutes a researcher configuration.\arxivfilltight

\paragraph{Target models and initial harnesses.}
\label{sec:target-models-harnesses}

All target agents begin with either Qwen3.5-9B-Base or MiniCPM5-2B-Base paired with a domain-specific initial harness. SWE-bench Verified and TB-Lite start from early versions of SWE-agent and Terminus, respectively. AppWorld uses a minimal code-generation harness, Telecom a lightweight text-based ReAct agent, and BFCL its official execution loop with a thin policy layer.

\paragraph{Execution settings and baselines.}
\label{sec:execution-baselines}

All runs follow the resource budget and development protocol in Section~\ref{sec:benchmark-design}. For each target model, the initial-system baseline (Base) pairs the base weights with each domain's initial harness, and the Reference configuration pairs its instruction-tuned counterpart with a domain-specific reference harness. We additionally evaluate the base model with the final harness to measure the gain from weight updates under the same harness. We report mean test performance across three independent development runs per researcher--benchmark pair and improvement over the initial system.\arxivfill

%% file: tables/table1_content.tex
\begingroup
\definecolor{BaselineBlue}{HTML}{E3F5FB}
\definecolor{EvoGainGreen}{RGB}{0,140,45}
\definecolor{EvoLossRed}{RGB}{190,20,32}
\definecolor{EvoNeutralGray}{RGB}{110,110,110}
\newlength{\EvoDeltaWidth}
\settowidth{\EvoDeltaWidth}{{\scriptsize +00.00}}

\newcommand{\gain}[1]{\,\makebox[\EvoDeltaWidth][l]{\textcolor{EvoGainGreen}{\scriptsize +#1}}}
\newcommand{\loss}[1]{\,\makebox[\EvoDeltaWidth][l]{\textcolor{EvoLossRed}{\scriptsize -#1}}}
\newcommand{\same}{\,\makebox[\EvoDeltaWidth][l]{\textcolor{EvoNeutralGray}{\scriptsize $\pm$0.00}}}
\newcommand{\nodelta}{%
  \,\makebox[\EvoDeltaWidth][l]{}%
}

\begin{table*}[t]
\centering
\caption{\textbf{Main results on CoEvoBench.} Target agents start from Qwen3.5-9B-Base (upper block) or MiniCPM5-2B-Base (lower block), each paired with a domain-specific initial harness. Researchers can modify model weights and the harness within a 24-hour development budget using four NVIDIA H20 GPUs. Results report mean test performance (\%) over three independent development runs per researcher--benchmark pair. In the lower block, Avg.\ uses weights from MiniCPM5-2B-Base's own Base and reference scores.\arxivfilltight}
\label{tab:main-leaderboard}
\small
\resizebox{\textwidth}{!}{%
\begin{tabular}{l cccccc}
\toprule
\textbf{\rule{0pt}{2ex}Researcher}
& \textbf{Avg.}\nodelta
& \shortstack{\textbf{SWE-bench}\\\textbf{Verified}}\nodelta
& \textbf{TB-Lite}\nodelta
& \textbf{AppWorld}\nodelta
& \shortstack{\textbf{$\tau^2$-bench}\\\textbf{Telecom}}\nodelta
& \textbf{BFCL-V4}\nodelta \\
\midrule
\rowcolor{BaselineBlue}
\multicolumn{7}{c}{\textit{Target model: Qwen3.5-9B-Base}} \\
\midrule
Base
  & 1.09\nodelta
  & 0.40\nodelta
  & 2.50\nodelta
  & 0.00\nodelta
  & 0.56\nodelta
  & 1.83\nodelta \\
\midrule
Kimi K3
  & 7.93\gain{6.84}
  & 5.67\gain{5.27}
  & 8.09\gain{5.59}
  & 3.27\gain{3.27}
  & 14.44\gain{13.88}
  & 11.63\gain{9.80} \\
Qwen3.8-max
  & 9.40\gain{8.31}
  & 11.46\gain{11.06}
  & 18.60\gain{16.10}
  & 5.23\gain{5.23}
  & 13.33\gain{12.77}
  & 4.51\gain{2.68} \\
GLM-5.3
  & 12.24\gain{11.15}
  & 16.07\gain{15.67}
  & 19.31\gain{16.81}
  & 12.42\gain{12.42}
  & 19.44\gain{18.88}
  & 2.94\gain{1.11} \\
Claude Opus 5
  & \underline{24.19}\gain{23.10}
  & \textbf{24.11}\gain{23.71}
  & \underline{19.67}\gain{17.17}
  & \underline{36.60}\gain{36.60}
  & \underline{36.11}\gain{35.55}
  & \underline{14.49}\gain{12.66} \\
GPT-6 Astra
  & \textbf{31.23}\gain{30.14}
  & \underline{23.58}\gain{23.18}
  & \textbf{24.43}\gain{21.93}
  & \textbf{37.91}\gain{37.91}
  & \textbf{73.89}\gain{73.33}
  & \textbf{25.88}\gain{24.05} \\
\midrule
\shortstack[l]{Instruct +\\Reference Harness}
  & 37.42\nodelta
  & 29.25\nodelta
  & 46.48\nodelta
  & 33.33\nodelta
  & 98.89\nodelta
  & 26.79\nodelta \\
\midrule
\rowcolor{BaselineBlue}
\multicolumn{7}{c}{\textit{Target model: MiniCPM5-2B-Base}} \\
\midrule
Base
  & 0.48\nodelta
  & 0.00\nodelta
  & 0.00\nodelta
  & 0.00\nodelta
  & 3.89\nodelta
  & 1.50\nodelta \\
\midrule
GLM-5.3
  & 5.76\gain{5.28}
  & 0.26\gain{0.26}
  & 1.35\gain{1.35}
  & 6.54\gain{6.54}
  & 16.30\gain{12.41}
  & 12.53\gain{11.03} \\
Kimi K3
  & 9.70\gain{9.22}
  & 0.79\gain{0.79}
  & \underline{1.67}\gain{1.67}
  & 11.76\gain{11.76}
  & 42.96\gain{39.07}
  & 14.40\gain{12.90} \\
Qwen3.8-max
  & 12.02\gain{11.54}
  & 3.95\gain{3.95}
  & 1.15\gain{1.15}
  & 13.73\gain{13.73}
  & 64.63\gain{60.74}
  & 14.21\gain{12.71} \\
Claude Opus 5
  & \underline{16.44}\gain{15.96}
  & \underline{5.40}\gain{5.40}
  & 1.24\gain{1.24}
  & \underline{20.26}\gain{20.26}
  & \underline{81.85}\gain{77.96}
  & \underline{19.13}\gain{17.63} \\
GPT-6 Astra
  & \textbf{18.46}\gain{17.98}
  & \textbf{8.04}\gain{8.04}
  & \textbf{1.86}\gain{1.86}
  & \textbf{21.57}\gain{21.57}
  & \textbf{96.11}\gain{92.22}
  & \textbf{19.90}\gain{18.40} \\
\midrule
\shortstack[l]{Instruct +\\Reference Harness}
  & 28.66\nodelta
  & 36.76\nodelta
  & 21.31\nodelta
  & 15.69\nodelta
  & 98.33\nodelta
  & 36.02\nodelta \\
\bottomrule
\end{tabular}
}
\end{table*}
\endgroup

%% file: sections/analysis.tex
\section{Analysis}
\label{sec:analysis}

\subsection{Diminishing Returns in Recursive Improvement}
\label{sec:recursive-improvement-dynamics}

In recursive improvement, each candidate builds on the previous system, so gains should accumulate. Yet development trajectories show that most of the improvement arrives at once.

\begin{figure}[t]
    \centering
    \includegraphics[width=\linewidth]{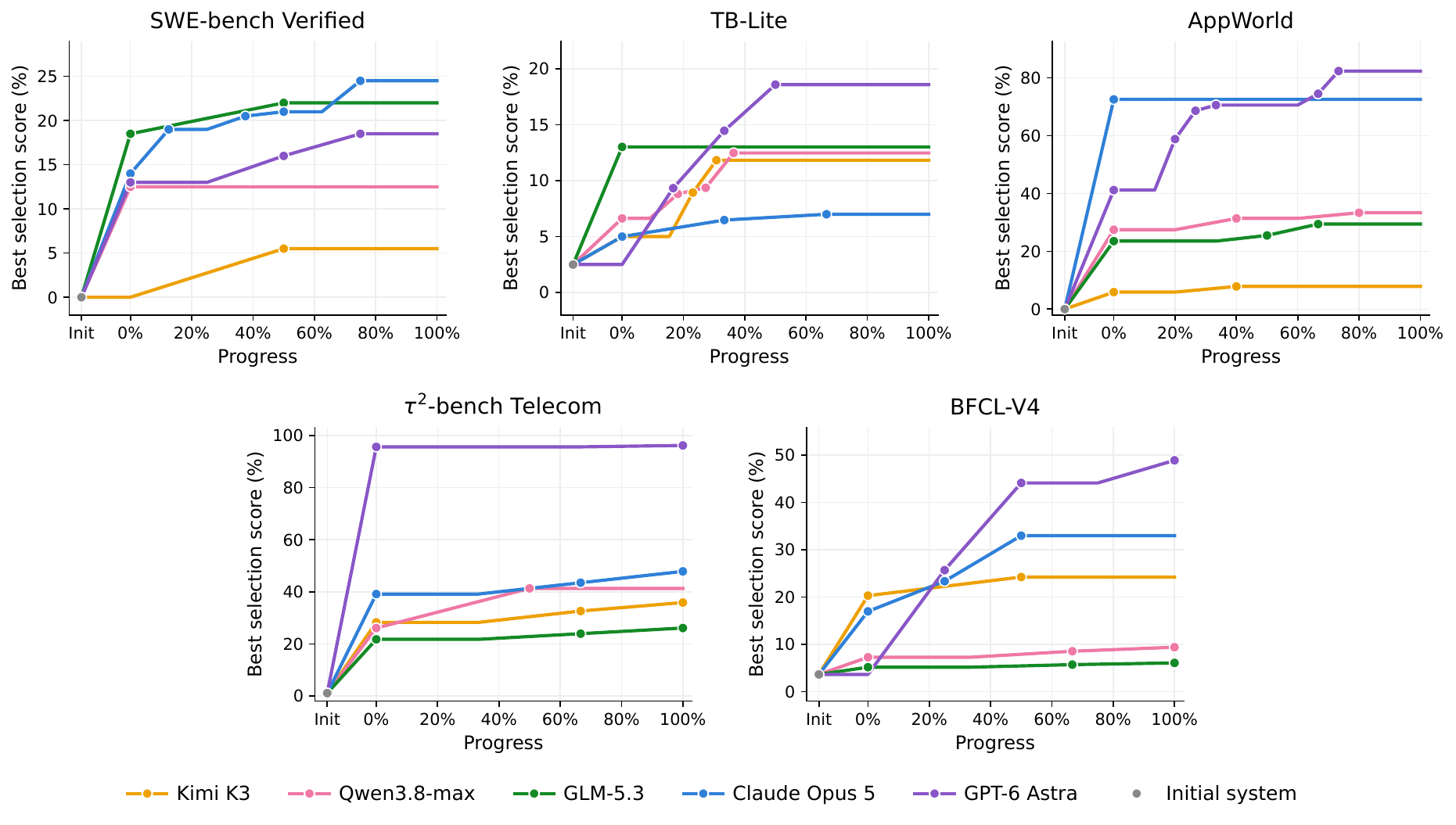}
    \caption{\textbf{Best-so-far development trajectories.} Each line follows one development run without audit findings per researcher--benchmark pair, the run with at least three evaluated candidates that sets the most new bests over the initial system (Init). Progress runs from the first evaluated candidate (0\%) to the last (100\%), so the rise from Init is the gain of initial development, and dots mark new bests. Among the 28 reviewed runs without audit findings that improve on the initial system, the first rise is the largest step in 25.\arxivfill}
    \label{fig:research-trajectories}
\end{figure}

\paragraph{Feedback-driven iteration adds limited gains.}
In most runs, the first candidate already captures most of the gain. The feedback-driven iterations that follow add comparatively little.
We split each run into initial development, which ends with the first candidate evaluated on the selection set, and feedback iteration, which carries that candidate to the best one.
In improving runs, the first candidate typically delivers about three quarters of the total gain and is usually the largest single step of the trajectory, a pattern that holds on all five benchmarks (Figure~\ref{fig:research-trajectories}).
The phase that recursion depends on, improving an already improved system, is the one that contributes least.

\paragraph{Deeper development yields fewer new gains.}
The further a run progresses, the harder each next gain becomes.
In runs with at least five candidates, the share of candidates that set a new best falls by more than half from the first half of feedback iteration to the second. To compare this pattern with chance, we shuffle the candidate order within each run, and only first-half candidates set new bests more often than in the shuffled runs.
The longest trajectory, an Astra AppWorld run with 16 candidates, shows the shape of this decline. Gains come quickly after the first candidate, and in the long stretch that follows only two candidates improve on the best.
Both late gains come from coordinated updates, in which the researcher first revises the harness and then retrains the model for it (Section~\ref{sec:model-harness-adaptation}).
What researchers lack is not the ability to improve a system once, but the ability to turn feedback from an already improved system into the next improvement.\arxivfill

\subsection{Model--Harness Co-Evolution}
\label{sec:model-harness-adaptation}

\input{tables/table_coupling}

Model and harness do not contribute independently. Which component should carry a gain depends on the task, and every change to one alters the value of the other. Of the 75 runs on Qwen3.5-9B-Base, 73 both revise the harness and launch training, and 54 submit trained weights. Table~\ref{tab:coupling}a shows that training carries over 80\% of the gain on BFCL-V4, while the harness carries most of the gain on the other four tasks, where the same harness with Qwen3.5-9B-Instruct still scores 6 to 33 points higher.

\paragraph{Frontier researchers struggle to train for long-horizon coding.}
On coding tasks, the strongest systems come mainly from the harness, and training has yet to pay off.
Harness revisions, spanning control flow, output parsing, prompts, decoding, and tool interfaces, let Base apply capabilities it already has, raising SWE-bench Verified and TB-Lite by 12.4 and 13.8 points (Table~\ref{tab:coupling}a). The same harness with Instruct scores another 6.1 and 15.8 points above the researchers' trained systems, so the model-side headroom is there.
Astra, the strongest researcher, trains in every coding run, yet each trained checkpoint falls below Base on the selection set, and it submits Base all six times. In one TB-Lite run, training halves the selection score under the same harness. Every researcher relies on offline fine-tuning, and none uses reinforcement learning.
Long-horizon coding demands end-to-end behavioral consistency, and today's frontier researchers rarely manage to train it into the target model.

\paragraph{Harness and training advance in alternation on interaction tasks.}
On interaction tasks, the two components divide the work. The harness makes the environment usable, for example through authentication, pagination, and interaction-history management, and training then teaches the model to act well within it.
Training adds gains on all three interaction tasks, 4.1 points on AppWorld, 1.6 on Telecom, and 10.1 on BFCL-V4, where the trained systems beat Instruct with the same harness by 4.4 points (Table~\ref{tab:coupling}a). The strongest researcher's choices show the contrast with coding. Astra submits trained weights in 7 of its 9 interaction runs, which beat Base under the same harness by 13.0 points on average (Table~\ref{tab:decisions}).
In an Opus AppWorld run, harness revisions raise test success from 0/51 to 3/51, and training then raises it to 12/51, three times the harness gain (Appendix~\ref{app:component-comparisons}).
Short, structured action sequences turn successful trajectories into direct learning signals, so on these tasks training pays off once the harness works.\arxivfill

\begin{figure}[t]
    \centering
    \includegraphics[width=\linewidth]{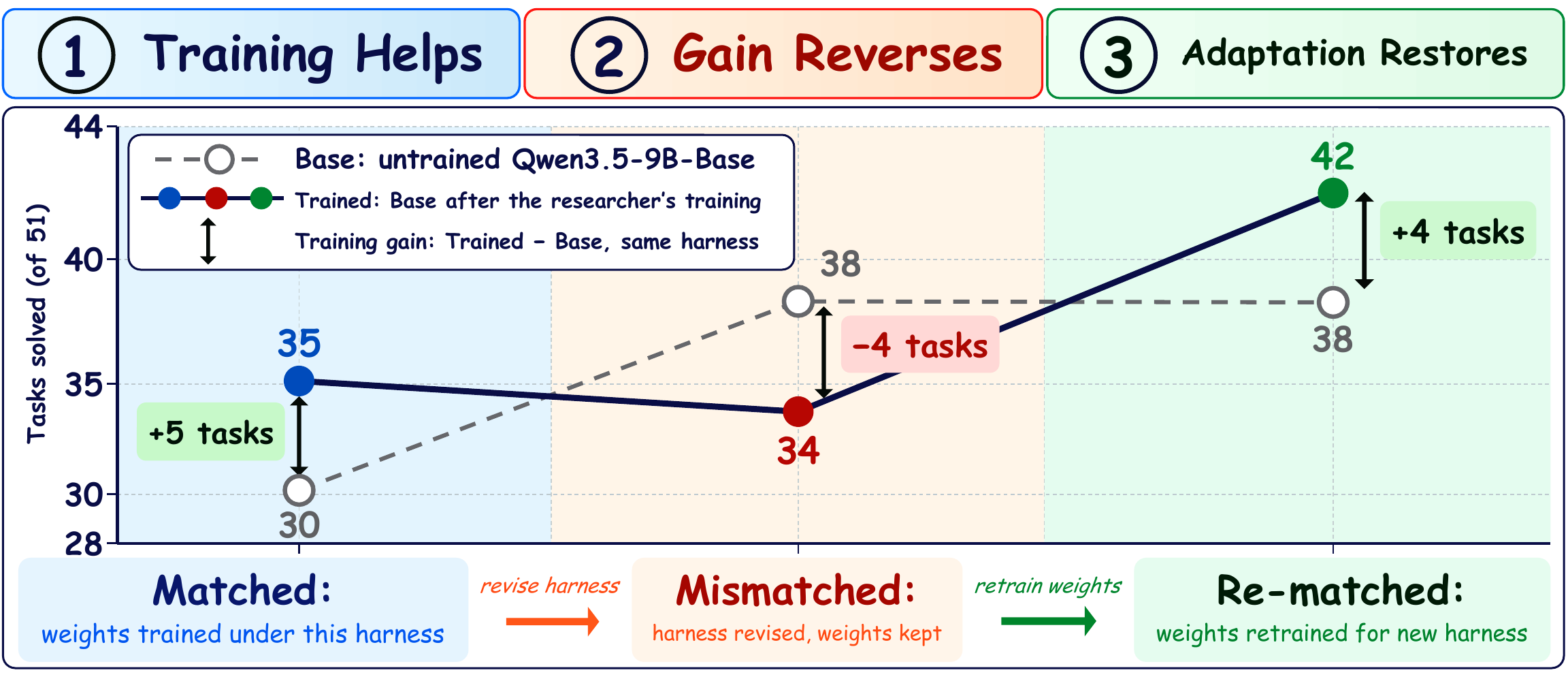}
    \caption{\textbf{A case study of model--harness co-evolution.} In an Astra development run on AppWorld, training helps under the earlier harness, its benefit reverses after a harness revision, and retraining on refreshed data restores the gain under the revised harness. Counts are tasks solved on the same 51-task selection set. Appendix~\ref{app:coupling} details each stage.\arxivfilltight}
    \label{fig:appworld-coupling}
\end{figure}

\paragraph{Harness revisions reshape the value of training.}
A harness revision can turn a helpful training update into a harmful one.
In an Astra AppWorld run, a training update first solves five more selection tasks than Base. After a harness revision, the same weights fall four tasks behind Base, and retraining on trajectories collected under the revised harness restores a gain of four tasks (Figure~\ref{fig:appworld-coupling}).
GLM shows the same dependence on AppWorld. Weights trained under an earlier harness score below Base, while weights retrained after a revision more than double Base's score (Appendix~\ref{app:component-comparisons}).
Training data are trajectories collected under one harness, and the behaviors learned from them fit that harness. A revision breaks the fit.
Cross-evaluating the 35 systems with trained weights in Table~\ref{tab:coupling}b shows that the coupling is general. The reference harness lifts Instruct from 30.2\% to 37.8\% but drops the trained weights from 19.8\% to 2.6\%. Every harness revision reopens the training decision.\arxivfill

Task characteristics thus decide whether current researchers can achieve co-evolution, and to what degree. On long-horizon coding, researchers rarely make training pay off, and improvement rests mainly on the harness. On short, structured interaction tasks, harness and training advance in alternation. Even where co-evolution succeeds, the two components stay coupled, and a change to either one resets the value of the other. Every further gain therefore requires deciding anew which component to change, based on what the current system reveals. Whether researchers can make this decision from their own results is the question of Section~\ref{sec:research-decisions}.\arxivfilltight

\subsection{Where the Loop Breaks}
\label{sec:research-decisions}

\input{tables/table_decisions}

Comparing the strongest researcher, GPT-6 Astra, with Qwen3.8-max, representative of the weaker researchers, shows where the research loop breaks. Table~\ref{tab:decisions} shows that stronger researchers revise the harness more and train less, while weaker researchers launch more training jobs and submit trained weights more often, yet gain far less from the weights they submit.
Qwen carries out every research step. In nearly every run, it diagnoses why the initial system fails, revises the harness, trains the model, and compares weights under the same harness.
Yet these results do not shape the three decisions that matter: which direction to take, when to abandon it, and which system counts as better.\arxivfill

\paragraph{Training is chosen before it is tested.}
Qwen decides to train before checking whether training is needed.
On SWE-bench Verified, both researchers find that Base cannot follow the required action format.
Astra first asks what the system can do without training. An interface rewritten to fit Base shows that the harness alone goes a long way, and when a brief LoRA trial falls below Base, Astra drops training.
Qwen first asks how to train. It settles on fine-tuning before testing it, rebuilds the harness around public training trajectories, and compares against Base only late in the run.
On coding tasks, where the harness carries the gains, this unexamined choice is the wrong one, and because gains are front-loaded, it costs most of the run.\arxivfill

\paragraph{Negative results change details, not the route.}
When results go against its route, Qwen changes the details, never the route.
On TB-Lite, its fine-tuned model shows no gain over Base. Qwen adds loop detection, generates a larger training set, and trains again, each time attributing the stall to insufficient data.
It uses results to swap checkpoints and adjust data, but never to question the decision to train.
Astra, after a single comparison in which training lowers the score, abandons training and submits Base with its harness.
The two runs end at nearly the same test score, and scaling up training changes nothing.
Under a fixed budget, the most valuable conclusion a researcher can reach is that a route does not work, and Qwen never draws it.\arxivfill

\paragraph{Candidates are judged on the tasks they were trained on.}
Qwen judges its candidates on the visible tasks, the same tasks its training data come from, so it cannot tell a real improvement from memorization.
This is not a submission formality.
Every round of recursive improvement rests on deciding whether the new system is better than the old one. Judged on data the model has seen, iteration accumulates overfitting, and the nomination, the only scored decision, makes the error final.
Qwen often picks its submission by visible-task scores, and when two checkpoints tie on the selection set, it lets the visible set decide.
Astra treats the selection set as a measuring instrument. Before submitting, it re-evaluates its finalists on the full selection set, overturns a smaller subset's ranking, and picks its strongest SWE-bench Verified system.
Judging improvement on tasks the model was not trained on lets iteration accumulate gains. This is what makes a loop recursive.

Astra lets each result decide the next step, which is why its first jump comes early and large.
Yet as Section~\ref{sec:recursive-improvement-dynamics} shows, even Astra finds it increasingly hard to extract the next step from an improved system.
Turning results into decisions is where recursive improvement stalls, for the weaker researcher from the start and for the strongest one after its first jump.\arxivfill

%% file: tables/table_coupling.tex
\begingroup
\definecolor{BaselineBlue}{HTML}{E3F5FB}
\definecolor{EvoGainGreen}{RGB}{0,140,45}
\definecolor{EvoLossRed}{RGB}{190,20,32}
\newlength{\CpDeltaWidth}
\settowidth{\CpDeltaWidth}{{\scriptsize +00.0}}
\newcommand{\cv}[1]{\makebox[1.9em][r]{#1}}
\newcommand{\cg}[1]{\,\makebox[\CpDeltaWidth][l]{\textcolor{EvoGainGreen}{\scriptsize +#1}}}
\newcommand{\cl}[1]{\,\makebox[\CpDeltaWidth][l]{\textcolor{EvoLossRed}{\scriptsize $-$#1}}}
\newcommand{\cn}{\,\makebox[\CpDeltaWidth][l]{}}
\begin{table}[t]
\centering
\caption{\textbf{Contributions of harness and training, and cross-evaluation of weights and harnesses.} Mean Test performance (\%) on Qwen3.5-9B. Harness is the harness evolved in each run. (a) The 53 submitted systems without an integrity issue. + Training uses the submitted weights, and + Instruct replaces them with Qwen3.5-9B-Instruct, which was never trained on any researcher harness. (b) The 35 of these systems that submit trained weights. Deltas are relative to the row above, and All averages every system in the panel.\arxivfilltight}
\label{tab:coupling}
\small
\setlength{\tabcolsep}{6pt}
\renewcommand{\arraystretch}{1.15}
\resizebox{\textwidth}{!}{%
\begin{tabular}{ll cc ccc c}
\toprule
 & & \multicolumn{2}{c}{\textbf{Coding}} & \multicolumn{3}{c}{\textbf{Interaction}} & \\
\cmidrule(lr){3-4}\cmidrule(lr){5-7}
 & & \shortstack{\textbf{SWE-bench}\\\textbf{Verified}}\cn & \textbf{TB-Lite}\cn & \textbf{AppWorld}\cn & \shortstack{\textbf{$\tau^2$-bench}\\\textbf{Telecom}}\cn & \textbf{BFCL-V4}\cn & \textbf{All}\cn \\
\midrule
\rowcolor{BaselineBlue}
\multicolumn{8}{c}{\textit{(a) Decomposition of the gain}} \\
\midrule
\multicolumn{2}{l}{Base} & \cv{0.4}\cn & \cv{2.5}\cn & \cv{0.0}\cn & \cv{0.6}\cn & \cv{1.8}\cn & \cv{1.2}\cn \\
\multicolumn{2}{l}{+ Harness} & \cv{12.8}\cg{12.4} & \cv{16.3}\cg{13.8} & \cv{21.8}\cg{21.8} & \cv{43.5}\cg{42.9} & \cv{4.0}\cg{2.2} & \cv{16.9}\cg{15.7} \\
\multicolumn{2}{l}{+ Harness + Training} & \cv{18.8}\cg{6.0} & \cv{17.6}\cg{1.3} & \cv{25.9}\cg{4.1} & \cv{45.1}\cg{1.6} & \cv{14.1}\cg{10.1} & \cv{21.8}\cg{4.9} \\
\multicolumn{2}{l}{+ Harness + Instruct} & \cv{24.9}\cg{6.1} & \cv{33.4}\cg{15.8} & \cv{47.5}\cg{21.6} & \cv{78.3}\cg{33.2} & \cv{9.7}\cl{4.4} & \cv{33.9}\cg{12.1} \\
\midrule
\rowcolor{BaselineBlue}
\multicolumn{8}{c}{\textit{(b) Cross-evaluation of systems with trained weights}} \\
\midrule
\multirow{2}{*}{Instruct} & Researcher harness & \cv{20.9}\cn & \cv{34.3}\cn & \cv{48.7}\cn & \cv{80.0}\cn & \cv{8.9}\cn & \cv{30.2}\cn \\
 & Reference harness & \cv{29.3}\cg{8.4} & \cv{46.5}\cg{12.2} & \cv{33.3}\cl{15.4} & \cv{98.9}\cg{18.9} & \cv{26.8}\cg{17.9} & \cv{37.8}\cg{7.6} \\
\midrule
\multirow{2}{*}{Trained} & Researcher harness & \cv{16.0}\cn & \cv{14.6}\cn & \cv{24.1}\cn & \cv{62.8}\cn & \cv{15.7}\cn & \cv{19.8}\cn \\
 & Reference harness & \cv{0.5}\cl{15.5} & \cv{2.2}\cl{12.4} & \cv{2.9}\cl{21.2} & \cv{8.0}\cl{54.8} & \cv{3.4}\cl{12.3} & \cv{2.6}\cl{17.2} \\
\bottomrule
\end{tabular}
}
\end{table}
\endgroup

%% file: tables/table_decisions.tex
\begin{table}[t]
\centering
\caption{\textbf{Research behavior of each researcher on Qwen3.5-9B.} Avg.\ is the weighted Test performance from Table~\ref{tab:main-leaderboard}. Per-run columns average the 15 development runs of each researcher. Trained submissions counts runs that submit trained weights, and training gain is the mean Test gain of these weights over Base weights under the same harness, over submissions without an integrity issue.}
\label{tab:decisions}
\small
\newcommand{\dv}[1]{\makebox[1.9em][r]{#1}}
\renewcommand{\arraystretch}{1.15}
\begin{tabular*}{\textwidth}{@{\extracolsep{\fill}}l c cc cc@{}}
\toprule
 & & \multicolumn{2}{c}{\textbf{Per run}} & & \\
\cmidrule(lr){3-4}
\textbf{Researcher} & \textbf{Avg.\ (\%)} & \textbf{Harness edits} & \textbf{Training jobs} & \textbf{Trained submissions} & \textbf{Training gain} \\
\midrule
Kimi K3 & \dv{7.93} & \dv{10.4} & \dv{6.9} & 14 / 15 & \dv{2.9} \\
Qwen3.8-max & \dv{9.40} & \dv{24.4} & \dv{3.9} & 11 / 15 & \dv{1.1} \\
GLM-5.3 & \dv{12.24} & \dv{21.8} & \dv{3.4} & 13 / 15 & \dv{7.8} \\
Claude Opus 5 & \dv{24.19} & \dv{29.9} & \dv{2.7} & \phantom{0}9 / 15 & \dv{12.3} \\
GPT-6 Astra & \dv{31.23} & \dv{28.1} & \dv{2.0} & \phantom{0}7 / 15 & \dv{13.0} \\
\bottomrule
\end{tabular*}
\end{table}

%% file: sections/conclusion.tex
\section{Conclusion}
\label{sec:conclusion}

We introduced CoEvoBench, a benchmark that places LLM researchers in the agent developer's position and asks whether they can drive the recursive co-evolution of model and harness across five agentic domains. The strongest researcher closes more than four fifths of the gap to an instruction-tuned reference system, but whether co-evolution happens, and to what degree, depends on the task. On long-horizon coding, the strongest systems rely mainly on the harness, as today's frontier researchers train only with offline fine-tuning and cannot yet make training pay off. On interaction tasks, harness and training advance in alternation, and each harness revision changes what training is worth. Researchers differ most in judgment. A weaker researcher completes every research step, yet its results rarely change its decisions, while the strongest lets each result decide the next step, though even it finds further gains harder after its first jump. Automating agent development, a concrete form of recursive self-improvement, depends on this judgment. Future work can extend CoEvoBench to more target models and to researchers that use reinforcement learning. We release the full protocol and the task splits, so that new researchers and specialized co-evolution pipelines can be evaluated under the same rules.\arxivfill

%% file: sections/appendix_related.tex
\section{Comparison with Related Work}
\label{app:related-comparison}

\DeclareRobustCommand{\cmk}{\textcolor[HTML]{2E8B57}{\ding{51}}}
\DeclareRobustCommand{\xmk}{\textcolor[HTML]{C0392B}{\ding{55}}}
\newsavebox{\pmkbox}
\sbox{\pmkbox}{\tikz[baseline=-0.55ex]{\draw[color={rgb,255:red,214;green,124;blue,26}, line width=0.5pt] (0,0) circle (0.6ex); \fill[color={rgb,255:red,214;green,124;blue,26}] (0,0) -- (90:0.6ex) arc (90:270:0.6ex) -- cycle;}}
\DeclareRobustCommand{\pmk}{\usebox{\pmkbox}}

Table~\ref{tab:related-comparison} compares CoEvoBench with benchmarks for automated AI research, harness optimization methods, and model--harness co-evolution methods. Benchmarks for automated AI research open only the model or only the delivered agent to change, harness optimization keeps the model fixed, and co-evolution methods update both components through a prescribed pipeline. CoEvoBench evaluates LLM researchers on agentic tasks with both components open and leaves the choice between them to the researcher.

\begin{table}[H]
\centering
\small
\setlength{\tabcolsep}{5pt}
\caption{\textbf{Comparison with related work.} Benchmark marks an evaluation of LLMs as researchers under a fixed protocol, Budget a fixed development budget for every researcher, Training updates to the target model's weights, Harness revisions to the code around the model, Agentic target tasks that require multi-turn interaction with tools or environments, and Choice that the researcher decides whether to train, revise the harness, or both. \cmk{} yes, \pmk{} partially, \xmk{} no.}
\label{tab:related-comparison}
\begin{tabular}{llcccccc}
\toprule
& \textbf{Work} & \textbf{Benchmark} & \textbf{Budget} & \textbf{Training} & \textbf{Harness} & \textbf{Agentic} & \textbf{Choice} \\
\midrule
\multirow{5}{*}{\shortstack[l]{Automated\\AI research}}
 & PostTrainBench~\citep{posttrainbench} & \cmk & \cmk & \cmk & \xmk & \pmk & \xmk \\
 & RSIBench-Data~\citep{rsibenchdata} & \cmk & \cmk & \cmk & \xmk & \cmk & \xmk \\
 & AI4AI-Bench~\citep{ai4aibench} & \cmk & \cmk & \cmk & \xmk & \xmk & \xmk \\
 & Agent$^2$ RL-Bench~\citep{agent2rlbench} & \cmk & \cmk & \cmk & \xmk & \cmk & \xmk \\
 & $\tau^\tau$-Bench~\citep{hypertau} & \cmk & \cmk & \xmk & \cmk & \cmk & \xmk \\
\midrule
\multirow{3}{*}{\shortstack[l]{Harness\\optimization}}
 & DGM~\citep{dgm} & \xmk & \xmk & \xmk & \cmk & \cmk & \xmk \\
 & Meta-Harness~\citep{metaharness} & \xmk & \xmk & \xmk & \cmk & \pmk & \xmk \\
 & DarwinX~\citep{darwinx} & \xmk & \xmk & \xmk & \cmk & \cmk & \xmk \\
\midrule
\multirow{5}{*}{\shortstack[l]{Model--harness\\co-evolution}}
 & SIA~\citep{sia} & \xmk & \xmk & \cmk & \cmk & \pmk & \cmk \\
 & Co-Harness~\citep{coharness} & \xmk & \xmk & \cmk & \cmk & \cmk & \xmk \\
 & HarnessForge~\citep{harnessforge} & \xmk & \xmk & \cmk & \cmk & \cmk & \xmk \\
 & HarnessX~\citep{harnessx} & \xmk & \xmk & \cmk & \cmk & \cmk & \xmk \\
 & ScienceBuddy~\citep{sciencebuddy} & \xmk & \xmk & \cmk & \cmk & \cmk & \xmk \\
\midrule
\rowcolor[HTML]{E3F5FB}
 & \textbf{CoEvoBench (ours)} & \cmk & \cmk & \cmk & \cmk & \cmk & \cmk \\
\bottomrule
\end{tabular}
\end{table}

%% file: sections/appendix.tex
\begingroup
\setlength{\textfloatsep}{10pt plus 2pt minus 2pt}
\setlength{\floatsep}{8pt plus 2pt minus 2pt}
\setlength{\intextsep}{9pt plus 2pt minus 2pt}
\renewcommand{\topfraction}{0.95}
\renewcommand{\bottomfraction}{0.9}
\renewcommand{\textfraction}{0.05}
\renewcommand{\floatpagefraction}{0.85}

\section{Researcher Workspace File Tree}
\label{app:workspace}

Each development run uses an independent workspace containing task instructions, an entry point to the initial model, the editable harness, and development-evaluation tools. In the tree below, \texttt{<benchmark>} denotes the run's benchmark: \texttt{swe}, \texttt{tb}, \texttt{appworld}, \texttt{tau}, or \texttt{bfcl}.

\begingroup\small
\begin{verbatim}
workspace/
|-- TASK.md                         Task and interface instructions
|-- base                            Initial model weights
|-- harness/<benchmark>/            Editable harness
|-- tasks_visible/<benchmark>/      Visible task material
|-- anonymous_pool_<benchmark>.txt  Anonymous evaluation task IDs
|-- eval                            Development evaluation entry
|-- timer.sh                        Research budget query
|-- model_path.txt                  Selected target model weights
`-- out/
    |-- ROUND.md                    Run-specific task and resources
    `-- <benchmark>_final/          Nominated final candidate
        |-- harness/                Candidate harness
        |-- model_path.txt          Candidate weights path
        |-- meta.json               Candidate metadata
        `-- serve_config.json       Model serving configuration
\end{verbatim}
\endgroup

The researcher edits \texttt{harness/}, trains and selects model weights, and calls \texttt{eval} to compare candidates. Training scripts, training material, and derived weights are created in the workspace. The final candidate directory is created on explicit nomination; candidate freezing follows Appendix~\ref{app:research-policy}.

\input{sections/appendix_results}

\section{Benchmark Tasks and Scoring}
\label{app:benchmarks}

CoEvoBench covers software repair, terminal operation, application interaction, telecom services, and function calling. Researchers improve target-agent performance by training the target model, modifying its harness, or combining both approaches.\arxivfill

\subsection{Task Pools and Split Construction}
\label{app:splits}

We partition tasks into visible development (Visible), selection (Selection), and final testing (Test) sets, with sizes in Table~\ref{tab:appendix-splits}. Fixed task manifests group or stratify tasks by domain structure.

\begin{table}[H]
\centering
\small
\setlength{\tabcolsep}{5pt}
\begin{tabular}{lrrrrp{.23\textwidth}}
\toprule
Benchmark & Visible & Selection & Test & Total & Grouping / stratification \\
\midrule
SWE-bench Verified & 40 & 200 & 253 & 493 & Repository \\
TB-Lite & 20 & 40 & 40 & 100 & Task difficulty \\
AppWorld easy & 27 & 51 & 51 & 129 & Scene / source split \\
$\tau^2$-bench Telecom & 23 & 46 & 45 & 114 & Read-only / write \\
BFCL-V4 & 235 & 470 & 560 & 1,265 & Task / memory groups \\
\bottomrule
\end{tabular}
\caption{\textbf{Frozen task partitions.} BFCL counts scoring records, including memory QA across backends. Each of the 45 Telecom Test tasks is evaluated four times, giving 180 trials. AppWorld groups by scene; BFCL groups by task family and memory scenario.\arxivfill}
\label{tab:appendix-splits}
\end{table}

\paragraph{SWE-bench Verified.}
The software-repair pool contains 493 tasks. Given the diversity of repository-level repair tasks and their binary success criterion, we use 200 tasks in Selection to provide a finer-grained evaluation signal for comparing candidate models and harnesses during development. Visible provides task statements; gold patches and private tests are managed by the evaluator. The target agent edits the repository; the verifier checks whether the issue is resolved.

\paragraph{TB-Lite.}
The terminal-operation pool contains 100 tasks, stratified by difficulty. Visible, Selection, and Test contain 20, 40, and 40 tasks, respectively. The agent interacts with the task environment through a terminal harness and receives the evaluator's task reward, including partial credit.

\paragraph{AppWorld easy.}
We select 129 difficulty-1 tasks from the normal and challenge collections, covering 43 scenes with three variants per scene. Variants of the same scene are assigned together, yielding 9 Visible scenes, 17 Selection scenes, and 17 Test scenes. Assignment is stratified by task source. The agent generates Python code that calls application APIs; task goal completion (TGC) measures complete-task success.

\paragraph{$\tau^2$-bench Telecom.}
We use the 114-task Telecom \emph{base} subset, stratified by read-only and write operations. Each of the 45 Test tasks is evaluated four times, giving 180 trials. The platform controls the user simulator and task environment; success requires passing all task checks.

\paragraph{BFCL-V4 multi-turn and memory.}
The pool contains 800 multi-turn records, with 200 in each of four categories, and 465 memory QA records, with 155 for each of three backends. Multi-turn variants in the same task family are assigned together, yielding 160/320/320 records. Each memory scenario is grouped with its prerequisite conversations and QA records across backends, yielding 75/150/240 records. Each memory QA evaluation independently replays its prerequisite conversation; accuracy is calculated per QA record.

\subsection{Metric Definitions}
\label{app:metrics}

All scores are displayed as percentages. For SWE-bench Verified and AppWorld, let $y_i\in\{0,1\}$ indicate complete-task success; the score is $100\sum_i y_i/N$. TB-Lite uses $100\sum_i r_i/N$, where $r_i\in[0,1]$ is the task reward.\arxivfill

For Telecom, let $y_{ik}$ denote success on task $i$ in trial $k$. We report mean single-trial success:
\begin{equation}
 s_{\mathrm{Telecom}}=\frac{100}{45\times4}\sum_{i=1}^{45}\sum_{k=1}^{4}y_{ik}.
\end{equation}

For BFCL, let $a_c$ be accuracy in multi-turn category $c$ and $m_d$ accuracy for memory backend $d$, each on the unit interval. The primary score is
\begin{equation}
 s_{\mathrm{BFCL}}=100\left(0.6\,\frac{1}{4}\sum_{c=1}^{4}a_c
 +0.4\,\frac{1}{3}\sum_{d=1}^{3}m_d\right).
\end{equation}
Test has equal record counts within each component, so the expression equals $100(0.6C_{\mathrm{MT}}/320+0.4C_{\mathrm{Mem}}/240)$, where $C$ denotes the number of successful records in the corresponding component.

\section{Researcher and Target-Agent Configuration}
\label{app:configuration}

\subsection{Researcher Configurations and Resources}

We use GPT-6 Astra, GLM-5.3, Qwen3.8-max, Kimi K3, and Claude Opus 5, all with high reasoning effort. Astra runs through Codex, while the other models run through Claude Code. Each researcher configuration comprises the model and its command-line tool.

Each development run receives a budget of 24 hours and four NVIDIA H20 GPUs, with early stopping permitted. Training, rollout generation, and model serving share these resources. The launcher sets the deadline, and the researcher can query the remaining budget.

\subsection{Initial Target Agents}
\label{app:seed-harnesses}

Target agents start from Qwen3.5-9B-Base. Researchers may retain the original weights or train derivatives using permitted data. Qwen3.5-9B-Instruct is used to construct the reference systems. Harness improvements include prompts, tool interaction, memory, and execution loops, relating to the harness optimization studied in Evo-Bench~\citep{evobench}.\arxivfill

\paragraph{Software repair and terminal operation.}
SWE-bench Verified starts from the SWE-agent agent--computer interface, while TB-Lite starts from the terminal-oriented Terminus harness. Both connect to the evaluator through the \texttt{minisweagent} interface, exposing action parsing, tool interaction, observation formatting, context management, and the execution loop. The initial SWE-bench Verified configuration uses strict action parsing, truncated observations, and a context limit. The selected local model generates target-agent responses.

\paragraph{AppWorld.}
The initial harness provides a minimal code-generation protocol. Researchers may modify agent construction and next-code generation, while the application environment and scorer remain fixed. The reference uses ReAct with full demonstrations and solving instructions.

\paragraph{BFCL.}
The initial harness combines a fixed execution loop with editable prompting and history-selection policies. It retains one ordinary user turn and uses replacement-style memory notes. Researchers may modify turn initialization and message selection; tool execution, environment transitions, and scoring remain fixed.

\paragraph{Telecom.}
The initial harness uses text-based ReAct and generates strict JSON actions. Native tool-schema injection and automatic action repair are disabled. GPT-4o simulates users (temperature 0, 600-token output limit); GPT-4.1 judges natural-language assertions and environment interfaces. Both follow the official $\tau^2$-bench evaluation configuration and are pinned to the dated snapshots \texttt{gpt-4o-2024-11-20} and \texttt{gpt-4.1-2025-04-14} at temperature 0.

\begin{table}[H]
\centering\small
\setlength{\tabcolsep}{4pt}
\renewcommand{\arraystretch}{1.12}
\begin{tabular}{@{}p{.17\textwidth}p{.40\textwidth}p{.38\textwidth}@{}}
\toprule
Benchmark & Weak initial harness & Stronger reference harness \\
\midrule
SWE-bench Verified & SWE-agent interface; strict parsing, truncated observations, and capped context. & mini-swe reference configuration. \\
TB-Lite & Terminal-oriented Terminus seed. & mini-swe reference configuration. \\
AppWorld & Minimal code-generation protocol, without full demonstrations or solving strategies. & ReAct with the reference demonstrations and solving prompt. \\
Telecom & Text ReAct with strict JSON actions; no native tool schema or automatic repair. & Telecom reference integration. \\
BFCL-V4 & Fixed execution loop; one ordinary user turn and replacement-style memory notes. & BFCL reference integration and policy. \\
\bottomrule
\end{tabular}
\caption{\textbf{Initial and reference harnesses used in our experiments.} The initial and stronger reference systems use Qwen3.5-9B-Base and Qwen3.5-9B-Instruct, respectively.}
\label{tab:appendix-harness-comparison}
\end{table}

The initial prompts of three harnesses are excerpted below from the released repository. Researchers may rewrite these prompts.

\begin{promptbox}{Initial SWE-bench Verified harness prompt (SWE-agent interface)}
SETTING: You are an autonomous programmer, and you're working directly in the command line with a special interface.

The special interface consists of a file editor that shows you {WINDOW} lines of a file at a time.
In addition to typical bash commands, you can also use the following commands to help you navigate and edit files.

COMMANDS:
{command_docs}

Please note that THE EDIT COMMAND REQUIRES PROPER INDENTATION.

RESPONSE FORMAT:
Your shell prompt is formatted as follows:
(Open file: <path>) <cwd> $

You need to format your output using two fields; discussion and command.
Your output should always include _one_ discussion and _one_ command field EXACTLY as in the following example:
DISCUSSION
First I'll start by using ls to see what files are in the current directory.
```
ls -a
```

You should only include a *SINGLE* command in the command section and then wait for a response from the shell before continuing with more discussion and commands.
However, the environment does NOT support interactive session commands (e.g. python, vim), so please do not invoke them.
\end{promptbox}

\begin{promptbox}{Initial AppWorld harness prompt}
USER:
Complete the task autonomously using the AppWorld Python REPL.
Each turn, output one fenced Python code block starting with ```python and ending with ```.
The code is executed and its printed output is returned. Variables persist between turns.
Expression values are NOT displayed automatically: always use print(...) to see a return value.
A response of "Execution successful." alone means no value was printed; it is not the API result.

The global object apis is already provided in the REPL. Do not import apis or define it yourself.
Apps are accessed through apis.<app_name>.<api_name>(**arguments).
Discover the interface using:
- print(apis.api_docs.show_app_descriptions())
- print(apis.api_docs.show_api_descriptions(app_name="..."))
- print(apis.api_docs.show_api_doc(app_name="...", api_name="..."))

Use the provided app APIs to interact with the world. References to files mean the file_system app.
Python standard-library code is available, but OS-level file and process access is prohibited.
Only make changes requested by the task. Choose reasonable values when details are unspecified.

When finished, call apis.supervisor.complete_task(answer=<requested value>) for a question, or apis.supervisor.complete_task() for an action.
Return the requested value directly, without explanatory prose. You may submit status="fail" if unable to solve the task.

USER:
My name is: {{ main_user.first_name }} {{ main_user.last_name }}. My personal email is {{ main_user.email }} and phone number is {{ main_user.phone_number }}.
Task: {{ instruction }}
\end{promptbox}

\begin{promptbox}{Initial BFCL-V4 harness prompt}
Respond to the user using the available tools when needed.

You should only return the function calls in your response.

If you decide to invoke any of the function(s), you MUST put it in the format of [func_name1(params_name1=params_value1, params_name2=params_value2...), func_name2(params)]. You SHOULD NOT include any other text in the response.

A response without tool calls ends the current turn.

Here is a list of functions in json format that you can invoke.
{functions}
\end{promptbox}

\subsection{Execution Limits and Serving Configuration}
\label{app:execution-limits}

Table~\ref{tab:appendix-execution} lists the initial configurations used for each benchmark. Development evaluations also record the candidate weights, harness, and serving configuration.

\begin{table}[H]
\centering
\small
\setlength{\tabcolsep}{5pt}
\begin{tabular}{lrrrrr}
\toprule
Setting & \shortstack{SWE-bench\\Verified} & TB-Lite & AppWorld & Telecom & BFCL \\
\midrule
Context length & 32,768 & 32,768 & 65,536 & 32,768 & 65,536 \\
Output tokens / request & 5,120 & 5,120 & 3,000 & 1,024 & 4,096 \\
Steps / loop constant & 80 & 120 & 50 & 200 & 20 \\
Temperature & 0 & 0 & 0 & 0 & 0.001 requested \\
Task wall limit (s) & 900 & 600 & None added & 1,200 & None added \\
\bottomrule
\end{tabular}
\caption{\textbf{Initial configurations used in our experiments.} BFCL uses \texttt{count > limit} loop semantics; serving raises the requested temperature of 0.001 to 0.01. For AppWorld and BFCL, the integration adds no task-level wall limit.\arxivfill}
\label{tab:appendix-execution}
\end{table}

\subsection{Choice of Target Model and Starting Point}
\label{app:target-choice}

We fix a 9B open target model so that the community can reproduce the benchmark. A full development run on the 9B target, including full fine-tuning, fits within four GPUs and 24 hours, whereas a larger target would multiply training cost. Absolute scores should therefore be read against the same-scale reference, Qwen3.5-9B-Instruct with a reference harness, which anchors each domain in Table~\ref{tab:main-leaderboard}. Leaderboard results on these benchmarks come from much larger frontier models. Against this reference, Opus raises SWE-bench Verified from 0.40\% to 24.11\%, over 80\% of the reference (29.25\%), Astra and Opus surpass it on AppWorld, and Astra nearly matches it on BFCL-V4.

Starting from Base with a weak harness leaves measurable headroom on both the model and the harness sides. Starting from Instruct with the reference harness would already reach 98.89\% on Telecom and leave no post-training headroom to measure, whereas the best researcher mean on TB-Lite (24.43\%) remains far below the reference (46.48\%). The improvements also carry over to a stronger model. Paired with Instruct, the evolved harnesses surpass the reference harness on AppWorld (Table~\ref{tab:coupling}).

\section{Permitted Research and Researcher Instructions}
\label{app:research-policy}

\subsection{Editable Components and Data Boundary}

Researchers may modify weights, the harness, or both. Available operations include diagnosing Visible failures; adjusting prompts, parsers, memory, and context; constructing and filtering training data; generating rollouts; selecting training recipes; and comparing and revising candidates. Data construction and selection relate to data-centric research in RSIBench-Data~\citep{rsibenchdata}.

Visible tasks support debugging, trajectory generation, and training. Network access is enabled, with datasets that directly expose benchmark task data blocked. Researchers may independently acquire permitted public or synthetic training material and record its sources and transformations. Selection/Test solutions, private tests, and per-task trajectories are protected: researchers may not retrieve them or construct data targeted at those instances. The benchmark infrastructure runs every Visible task in a sandbox that returns its reward and supports over 1,000 concurrent rollouts, so researchers can train with reinforcement learning as well as supervised fine-tuning.\arxivfill

The fixed \texttt{qwen3.8-27b} teacher may generate permitted training data within a \$100 calling budget per development run; evaluation tasks are executed by the target student model. Research uses the supplied Base or its permitted derivatives within the assigned workspace and resources.

\subsection{Development Feedback and Candidate Freezing}

Researchers receive execution details on Visible and score feedback on Selection. Selection task statements and trajectories remain hidden. Development evaluations record task subsets and repetition counts; Test is reserved for final evaluation. Because researchers evaluate many candidates on the same Selection set and keep the best, Selection scores are optimistic estimates of Test performance. In Astra AppWorld Run 2, for example, the final candidate solves 42/51 Selection tasks and 22/51 Test tasks, still above the reference system's 33.33\%. Only the Test set, used once after nomination, measures generalization.\arxivfill

One model--harness pair is frozen at the end of development. The researcher nominates a candidate in \texttt{out/<benchmark>\_final/}; without an explicit nomination, the submission is the workspace harness and the weights selected by \texttt{model\_path.txt} at the deadline, defaulting to Base. The operator records candidate identities, supplies replay metadata, and runs Test. Ordinary final evaluation leaves the frozen weights and harness unchanged; weight reversion required by the integrity policy follows Appendix~\ref{app:integrity}.

\subsection{Researcher Prompts}
\label{app:prompt}

Each researcher receives the common task brief, a domain task brief, and a research-tools guide in its workspace, and the loop prompt starts and resumes every research session. The excerpts below come from the released repository.

\begin{promptbox}{Common task brief}
Given the supplied 9B Base model and the initial harness for a benchmark,
improve the target system within the assigned time and compute budget.
You may modify the weights, the harness, or both, and choose whether
to train.

Use Visible tasks and permitted evaluation feedback to diagnose
failures and compare candidates. You may acquire, generate, and
transform permitted training material. Do not retrieve protected
Selection/Test solutions, private tests, or trajectories, or
construct data targeted at those instances.

Use the supplied Base or its permitted derivatives. The fixed teacher
is for training-data generation; the student independently completes
evaluation tasks. Keep benchmark interfaces runnable and respect
scoring, timing, and resource boundaries.

Select and freeze a final model-harness pair for final evaluation by
the operator. You may stop development early.
\end{promptbox}

\begin{promptbox}{Researcher loop prompt}
Read TASK.md and RESEARCH_TOOLS.md and continue the research.
- Work under /workspace, preserving existing artifacts and notes.
- Recover your plan from workspace artifacts, not a corrupted conversation.
- Check ./timer.sh status. With less than 30 minutes left, finish the nomination.
- Evaluate candidates with ./eval (visible for diagnosis, selection
  for generalization).
- Keep deliverables updated under out/ per TASK.md.
- Before finishing, write out/DONE.md AND run ./submit. DONE.md alone
  does not complete a submission.
\end{promptbox}

\begin{promptbox}{SWE-bench Verified task brief (excerpt)}
# SWE-bench - model and harness co-evolution

You are the researcher for one SWE development run. Improve the supplied 9B
base model, its initial harness, or both within the assigned budget. Check
./timer.sh status; recovery retains the workspace and original deadline.

## Workspace and editable harness
- base/ is read-only.
- harness/src/minisweagent/agents/default.py adapts the initial SWE-agent ACI
  loop. Prompts and shell commands are in agents/aci/.
- Keep models/_transport.py unchanged and use the provided environment for
  evaluated commands. Other harness logic is editable.
- serve_config.json is the editable initial serving recipe.

## Evaluation and feedback
The frozen subset contains 40 Visible, 200 Selection and 253 Test instances.
Visible records provide scores and exit categories. Selection returns aggregate
feedback, and Test is operator-only.

  ./eval --split visible --model ./base --workers 4
  ./eval --split selection --model work/weights/v1 --workers 4
  ./eval --split selection --tasks A-xxxxxxxx --model work/weights/v1 --k 2

SWE scores the official tests against the candidate's real diff. Submitted is
not proof of resolution. Preserve negative results, settings and sample counts.

## Serving and training
Choose training materials independently of protected Selection/Test instances.
Permitted public repositories, other tasks, human-written solutions, custom
verifiers and assigned-model rollouts may be used. Do not retrieve protected
solutions, private tests or trajectories even if publicly downloadable.

  ./contamination check --input work/train.jsonl --manifest out/swe_final/train/corpus.json
  ./track -- python work/train.py

Preserve the base model architecture, tokenizer and processor assets. Merge LoRA
adapters into full weights for final replay.

## Nomination
out/swe_final/
  harness/
  model_path.txt      /workspace/work/weights/v1, or /base when unchanged
  meta.json           bench, changes, settings and results
  serve_config.json
  train/              recipe, data provenance and dependencies if trained

For trained models include a same-harness trained-vs-base control. Before the
deadline, assemble the nomination, write out/DONE.md and run ./submit --bench swe.
\end{promptbox}

\begin{promptbox}{Research tools guide (excerpt)}
You may independently acquire, generate and transform training data, use other
training tasks, write verifiers and construct task environments. Provided datasets
and images are starting resources, not a whitelist. Selection/Test restrictions
still apply.

## Downloading data and code
Use hf, curl, Python and Git freely for permitted training materials. Record
source URLs, revisions, checksums and data transformations in your training recipe.

## Task environments and interactive sandboxes
  ./sandbox create --from-dir work/my-task --cpu 2 --memory 4 --ttl 1800
  ./sandbox exec SESSION_ID 'python /task/verify.py' --cwd /task --timeout 120
  ./sandbox upload SESSION_ID work/input.bin /task/input.bin
  ./sandbox export SESSION_ID work/environment.jsonl
  ./sandbox close SESSION_ID

You can write and run your own verifier, upload a third-party training task or
build synthetic tasks this way. Use Visible ./eval for evaluation feedback; do not
retrieve protected evaluation environments through development tools.

## Rollout and training
  from coevobench.research.client import Client
  client = Client()
  session = client.call("sandbox.create", image="python:3.11-slim", ttl_s=300)

You control the model, prompts, steps, sampling, retries, reward design and
training algorithm. Use ./track -- python work/train.py to time the actual
training process.
\end{promptbox}

\section{Result Aggregation and Independent Runs}
\label{app:reporting}

\subsection{Baselines and Cross-Domain Aggregation}

The Base reference combines the original weights with the initial harness. The stronger reference combines Qwen3.5-9B-Instruct with a reference harness: mini-swe for SWE-bench Verified and TB-Lite, ReAct for AppWorld, and the respective reference integrations for Telecom and BFCL.

For benchmark $b$, Base and stronger-reference scores $B_b$ and $I_b$ define fixed normalized weights:
\begin{equation}
w_b=\frac{(I_b-B_b)^{-1}}{\sum_j(I_j-B_j)^{-1}},\qquad
\mathrm{Avg.}=\sum_b w_b s_b.
\end{equation}
Here $s_b$ is the researcher model's three-run mean on benchmark $b$. All researcher models use the same weights, which depend only on the Base and stronger-reference scores (Table~\ref{tab:appendix-weights}).\arxivfill

\begin{table}[H]
\centering
\small
\setlength{\tabcolsep}{4pt}
\begin{tabular*}{\textwidth}{@{\extracolsep{\fill}}lccccc@{}}
\toprule
Benchmark & \shortstack{SWE-bench\\Verified} & TB-Lite & AppWorld & Telecom & BFCL-V4 \\
\midrule
Weight (\%) & 25.18 & 16.52 & 21.80 & 7.39 & 29.11 \\
\bottomrule
\end{tabular*}
\caption{\textbf{Fixed scoring weights.} Displayed weights are rounded; calculations use their full normalized precision. These constants weight the benchmark means of the runs retained in Table~\ref{tab:independent-run-results}.}
\label{tab:appendix-weights}
\end{table}

\input{sections/appendix_efficiency}

\section{Development Trajectories and Case Studies}
\label{app:analysis}

\subsection{Trajectory Records and Candidate Comparisons}
\label{app:trajectory-method}

Trajectories link candidate harnesses, model weights, training recipes, and evaluation feedback within each development run. The cases below correspond to specific runs in Table~\ref{tab:independent-run-results}, using Selection scores to analyze development decisions and reporting the final candidates' Test scores.

\subsection{AppWorld: Reversing and Restoring a Training Gain}
\label{app:coupling}

In Astra AppWorld Run 2, the researcher first improves code continuation, API-documentation discovery, authentication, pagination, and context management. Two complete examples from successful Visible trajectories form the \textbf{example-enhanced harness}. Initial LoRA training increases Selection success from 30/51 to 35/51 under this harness.

The researcher adds attachment authentication, recursive live-path lookup, and relationship-based contact filtering. First attempts use temperature zero, while repeated failures retain an escape strategy. This \textbf{tool-enhanced harness} raises Base to 38/51, while the initial trained weights achieve 34/51. The paired results in Table~\ref{tab:appendix-coupling} give a training-gain difference of \mbox{$(34-38)-(35-30)=-9$ tasks}.

\begin{table}[H]
\centering
\small
\setlength{\tabcolsep}{4pt}
\begin{tabular}{llrrr}
\toprule
Harness & Weights & \shortstack{Visible\\successes} & \shortstack{Selection\\successes} & \shortstack{Gain over\\same-harness Base} \\
\midrule
Example-enhanced & Base & 16/27 & 30/51 (58.82\%) & --- \\
Example-enhanced & Initial training & 17/27 & 35/51 (68.63\%) & $+5$ tasks \\
\midrule
Tool-enhanced & Base & 17/27 & 38/51 (74.51\%) & --- \\
Tool-enhanced & Initial training & 17/27 & 34/51 (66.67\%) & $-4$ tasks \\
Tool-enhanced & Expanded training & 20/27 & 42/51 (82.35\%) & $+4$ tasks \\
\bottomrule
\end{tabular}
\caption{\textbf{Model--harness comparisons in Astra AppWorld Run 2.} Each Selection result evaluates the same 51 tasks once. Trained weights are first-epoch checkpoints; model comparisons within each harness version were checked against harness file fingerprints.\arxivfill}
\label{tab:appendix-coupling}
\end{table}

The researcher expands the successful-trajectory data and increases the maximum training sequence length from 16,384 to 32,768. The expanded training reaches 42/51 under the tool-enhanced harness, restoring a positive training gain. Table~\ref{tab:appendix-training} gives the two training configurations.

\begin{table}[H]
\centering
\small
\begin{tabular}{lrr}
\toprule
Training setting & Initial training & Expanded training \\
\midrule
Input / usable records & 99 / 91 & 189 / 183 \\
Maximum sequence length & 16,384 & 32,768 \\
LoRA rank / alpha & 16 / 32 & 16 / 32 \\
Learning rate & $8\times10^{-5}$ & $8\times10^{-5}$ \\
Gradient accumulation & 4 & 4 \\
Checkpoint compared & Epoch 1 & Epoch 1 \\
\bottomrule
\end{tabular}
\caption{\textbf{Training configurations for the AppWorld coupling case.} Records exceeding the sequence-length limit are filtered. Expanded-training inputs cover 19 tasks. Both configurations use seed 100 and train for two epochs; Table~\ref{tab:appendix-coupling} compares first-epoch checkpoints.}
\label{tab:appendix-training}
\end{table}

The final candidate combines the tool-enhanced harness with the first-epoch checkpoint from expanded training. The operator converts the serving configuration to the evaluator's format and adjusts the service port, preserving the nominated weights and harness. The candidate solves 22/51 Test tasks (43.14\%). The paired Selection comparisons trace the training gain from $+5$ to $-4$ solved tasks after the harness change, then to $+4$ after expanding the data and sequence length. Figure~\ref{fig:appendix-coupling-detail} summarizes the motivation, changes, and outcome of each stage.

\begin{figure}[H]
    \centering
    \includegraphics[width=\linewidth]{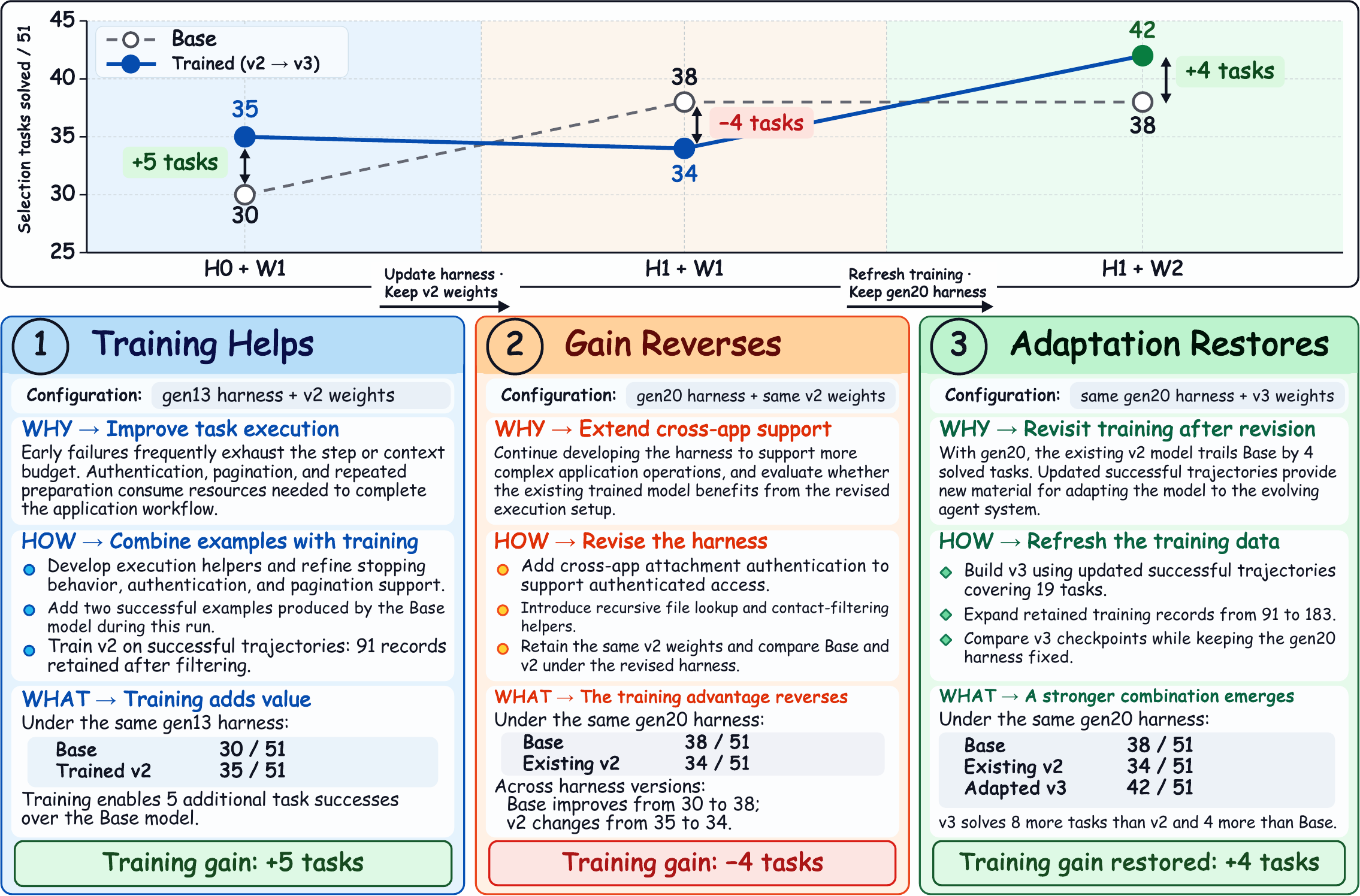}
    \caption{\textbf{Stages of the AppWorld coupling case.} The top panel compares the number of tasks solved by the base and trained models on the same 51-task selection set; the bottom panels summarize the motivation, changes, and outcomes at each stage.}
    \label{fig:appendix-coupling-detail}
\end{figure}

\subsection{AppWorld: Harness Optimization and Self-Distillation}
\label{app:plateau}

In Opus AppWorld Run 2, the researcher constructs a harness with initialization of time, user information, and credentials; login and pagination helpers; an API-name index; code-block stopping; empty-code and repeated-code resampling; and history compression. The fixed candidate is evaluated on Selection 35 times, yielding 1,159 successes out of 1,785 trials (64.93\%).

The researcher uses 90 successful trajectories for self-distillation. Two first-epoch training variants achieve 56.90\% (rank 8, learning rate $2\times10^{-5}$) and 47.70\% (rank 16, learning rate $5\times10^{-5}$). Removing the pre-submission self-check then yields 741 successes across 22 Selection evaluations, with 1,122 trials and a mean of 66.04\%. The final candidate uses this harness, which removes one generation step, with Base, achieving 49.02\% on Test.\arxivfill

The training-gain example in the main text, which improves Test success from 0/51 to 3/51 through harness changes and then to 12/51 through training, corresponds to Opus AppWorld Run 3 (23.53\%).

\subsection{Coding Tasks: Interface Adaptation and Training Regressions}
\label{app:coding-cases}

In Astra TB-Lite Run 1, training lowers Selection reward from Base's 8.59\% to 4.25\% under the same harness with the stopping configuration repaired; a Base few-shot variant achieves 5.44\%. Another harness with JSON recovery achieves 13.10\% with Base and becomes the final candidate, scoring 17.50\% on Test. This run also repairs checkpoint-serialization issues and ultimately uses Base with an improved harness.

GLM SWE-bench Verified Run 2 trains the model to adapt to an execution interface aligned with public successful trajectories. The trained checkpoint solves 44/200 Selection tasks, with a confirmation evaluation at 43/200, while Base solves 0/200 under the same interface. This case illustrates model adaptation to a specific execution interface.\arxivfill

Astra's three SWE-bench Verified Test scores are 19.76\%, 28.85\%, and 22.13\%; its three TB-Lite scores are 17.50\%, 31.03\%, and 24.76\%. All use Base with the harness developed in the corresponding run. The score ranges are 9.09 and 13.53 percentage points, respectively.

\section{Evaluation of Runs with Audit Violations}
\label{app:integrity}

For runs with a confirmed audit violation, final evaluation reverts the model weights to Qwen3.5-9B-Base, keeps the run's evolved harness, and measures this system on Test. Appendix~\ref{app:audit} describes the audit procedure and its findings.\arxivfill

\endgroup

%% file: sections/appendix_results.tex
\section{Individual Experiment Results}
\label{app:individual-results}

Table~\ref{tab:independent-run-results} reports three independent development runs per researcher--benchmark pair.

\input{tables/table2_content}

%% file: tables/table2_content.tex
\begingroup
\definecolor{RunGroupBlue}{HTML}{E3F5FB}
\begin{table}[H]
\centering
\caption{\textbf{Individual experiment results on the frozen Test partition (\%).} Three runs and their mean are shown for each benchmark. All researcher models use high reasoning effort. $\blacklozenge$ marks runs with a confirmed integrity issue (Appendix~\ref{app:audit}).}
\label{tab:independent-run-results}
\label{tab:appendix-run-scores}
\small
\setlength{\tabcolsep}{3pt}
\renewcommand{\arraystretch}{1.04}
\begin{tabular*}{\textwidth}{@{\extracolsep{\fill}}llccccc@{}}
\toprule
Benchmark & Run & Kimi K3 & Qwen3.8-max & GLM-5.3 & \shortstack{Claude\\Opus 5} & \shortstack{GPT-6\\Astra} \\
\midrule
SWE-bench Verified & Run 1 & 0.79\rlap{$^\blacklozenge$} & 17.79 & 4.35 & 24.11 & 19.76 \\
 & Run 2 & 0.79 & 0.40\rlap{$^\blacklozenge$} & 23.72 & 20.55 & 28.85 \\
 & Run 3 & 15.42\rlap{$^\blacklozenge$} & 16.21 & 20.16 & 27.67 & 22.13 \\
 & \textbf{Mean} & \textbf{5.67} & \textbf{11.46} & \textbf{16.07} & \textbf{24.11} & \textbf{23.58} \\
\midrule
TB-Lite & Run 1 & 6.00 & 20.03 & 23.47\rlap{$^\blacklozenge$} & 14.17 & 17.50 \\
 & Run 2 & 15.03 & 18.15\rlap{$^\blacklozenge$} & 16.00 & 20.68 & 31.03 \\
 & Run 3 & 3.25 & 17.62 & 18.47 & 24.17 & 24.76 \\
 & \textbf{Mean} & \textbf{8.09} & \textbf{18.60} & \textbf{19.31} & \textbf{19.67} & \textbf{24.43} \\
\midrule
AppWorld easy & Run 1 & 5.88\rlap{$^\blacklozenge$} & 7.84 & 5.88 & 37.25 & 37.25 \\
 & Run 2 & 1.96\rlap{$^\blacklozenge$} & 0.00\rlap{$^\blacklozenge$} & 13.73 & 49.02 & 43.14 \\
 & Run 3 & 1.96\rlap{$^\blacklozenge$} & 7.84 & 17.65\rlap{$^\blacklozenge$} & 23.53 & 33.33 \\
 & \textbf{Mean} & \textbf{3.27} & \textbf{5.23} & \textbf{12.42} & \textbf{36.60} & \textbf{37.91} \\
\midrule
$\tau^2$-bench Telecom & Run 1 & 17.78\rlap{$^\blacklozenge$} & 11.67\rlap{$^\blacklozenge$} & 18.33\rlap{$^\blacklozenge$} & 16.11 & 37.22 \\
 & Run 2 & 2.78\rlap{$^\blacklozenge$} & 18.33\rlap{$^\blacklozenge$} & 20.00\rlap{$^\blacklozenge$} & 17.78\rlap{$^\blacklozenge$} & 96.11 \\
 & Run 3 & 22.78 & 10.00 & 20.00\rlap{$^\blacklozenge$} & 74.44\rlap{$^\blacklozenge$} & 88.33 \\
 & \textbf{Mean} & \textbf{14.44} & \textbf{13.33} & \textbf{19.44} & \textbf{36.11} & \textbf{73.89} \\
\midrule
BFCL-V4 & Run 1 & 13.56 & 4.17\rlap{$^\blacklozenge$} & 2.00\rlap{$^\blacklozenge$} & 2.83 & 26.10 \\
 & Run 2 & 13.96 & 5.06 & 4.00 & 19.75 & 19.54 \\
 & Run 3 & 7.38 & 4.29 & 2.83\rlap{$^\blacklozenge$} & 20.88 & 31.98 \\
 & \textbf{Mean} & \textbf{11.63} & \textbf{4.51} & \textbf{2.94} & \textbf{14.49} & \textbf{25.88} \\
\bottomrule
\end{tabular*}
\end{table}
\endgroup

%% file: sections/appendix_efficiency.tex
\section{Per-Run Token Usage and API Cost}
\label{app:runtime}

Table~\ref{tab:appendix-cost} reports the token usage and API cost of a development run on Qwen3.5-9B-Base. Token counts include cache reads and writes. Researcher API cost applies each model's official list price to its own token usage, and teacher API cost is a range over two regional price schedules. Spending does not track performance. GLM has the lowest researcher API cost but a low score, Qwen has the highest, and Astra reaches the highest score with the lowest teacher cost.

\begin{table}[H]
\centering
\small
\setlength{\tabcolsep}{6pt}
\begin{tabular}{lrrrrr}
\toprule
 & \multicolumn{2}{c}{Tokens (M)} & \multicolumn{2}{c}{API cost} & \\
\cmidrule(lr){2-3}\cmidrule(lr){4-5}
Researcher & Input & Output & Researcher & Teacher & Avg.\ score \\
\midrule
Kimi K3 & 77.0 & 0.14 & \$49 & \$17--23 & 7.93 \\
Qwen3.8-max & 97.9 & 0.86 & \$160 & \$28--39 & 9.40 \\
GLM-5.3 & 60.6 & 0.25 & \$23 & \$24--32 & 12.24 \\
Claude Opus 5 & 42.5 & 0.25 & \$120 & \$44--58 & 24.19 \\
GPT-6 Astra & 48.0 & 0.22 & \$100 & \$2--3 & 31.23 \\
\midrule
Mean & 65.3 & 0.34 & \$91 & \$23--31 & --- \\
\bottomrule
\end{tabular}
\caption{\textbf{Per-run token usage and API cost of development.} Avg.\ score is the weighted Test performance (\%) from Table~\ref{tab:main-leaderboard}, and the other columns are per-run means. Researcher API cost uses official list prices.}
\label{tab:appendix-cost}
\end{table}

%% file: sections/appendix_component_comparisons.tex
\section{Model--Harness Updates Across Development Runs}
\label{app:component-comparisons}

We examine model and harness updates through comparisons of recorded system configurations. Table~\ref{tab:component-comparisons} combines final-system evaluations on AppWorld with development comparisons on SWE-bench Verified, TB-Lite, and AppWorld. Across the three AppWorld test comparisons, harness revisions improve the initial system, and updated weights provide further gains under the final harness. Development comparisons show both positive and negative training gains. In the TB-Lite run, averaging base and trained weights yields a higher selection score than either set of weights alone.

\begin{table}[H]
\centering
\caption{\textbf{Model and harness configurations across development runs.} Scores are percentages. Panel (a) reports AppWorld test scores with one trial per task. Panel (b) reports development evaluations; $n$ is the number of evaluated tasks. Run numbers follow Table~\ref{tab:independent-run-results}. Recorded harness versions are shown; SWE-bench Verified and TB-Lite compare named development configurations.}
\label{tab:component-comparisons}
\small
\begin{tabular}{lrrr}
\toprule
\multicolumn{4}{l}{\textbf{(a) AppWorld Test: initial and final components}}\\
Researcher / Run & Initial system & Base + final harness & Updated + final harness\\
\midrule
Opus / Run 1 & 0.00 & 19.61 & 37.25\\
Opus / Run 3 & 0.00 & 5.88 & 23.53\\
GLM / Run 2 & 0.00 & 7.84 & 13.73\\
\bottomrule
\end{tabular}

\vspace{6pt}
\begin{tabular}{lllr}
\toprule
\multicolumn{4}{l}{\textbf{(b) Development configurations}}\\
Benchmark / Run & Evaluation & Configuration & Score\\
\midrule
SWE-bench Verified & Selection, $n=40$ & Base & 2.50\\
Qwen / Run 1 & & SFT epoch2 & 12.50\\
\midrule
TB-Lite & Selection, $n=40$ & Base & 11.57\\
Qwen / Run 3 & & SFT-v3 & 9.08\\
& & $0.5\,\text{Base}+0.5\,\text{SFT-v3}$ & 16.83\\
\midrule
AppWorld & Visible, $n=27$ & Harness v2 + Base & 11.11\\
GLM / Run 2 & & Harness v2 + SFT-r1 & 7.41\\
& & Harness v4 + Base & 7.41\\
& & Harness v4 + SFT-r3 & 18.52\\
\bottomrule
\end{tabular}
\end{table}

The dynamic model--harness interaction in Figure~\ref{fig:appworld-coupling}, including paired configurations and training settings, is detailed in Appendix~\ref{app:coupling}.\arxivfill

%% file: sections/appendix_evaluation.tex
\section{Repeated Evaluation}
\label{app:repeat}

We evaluate each final system three times on Test on three benchmarks. Table~\ref{tab:appendix-repeat} compares the standard deviation of these repeated evaluations with the gaps between researcher means. Evaluation noise is far smaller than the differences between researchers, so the large variation across the three development runs of the same researcher comes from the research process itself.

\begin{table}[H]
\centering
\small
\begin{tabular}{lrr}
\toprule
Benchmark & Std.\ of repeated evaluations & Researcher means, best minus worst \\
\midrule
SWE-bench Verified & 1.0 & 18.4 \\
TB-Lite & 2.4 & 16.3 \\
BFCL-V4 & 0.4 & 22.9 \\
\bottomrule
\end{tabular}
\caption{\textbf{Evaluation noise of a frozen system versus differences between researchers} (points).}
\label{tab:appendix-repeat}
\end{table}

\section{Audit Procedure and Findings}
\label{app:audit}

\paragraph{Procedure.}
After each run, an agent applies the same audit skill to the complete development trajectory, including tool calls, file reads and writes, commands, training data lineage, and network logs during evaluation, and screens the whole trajectory regardless of its length. The skill defines six checks: obtaining protected Test or Selection material, substituting unauthorized weights, training on prohibited data, task-specific harness customization, reading files outside the workspace, and the integrity of the evaluation chain. Each check returns one of four verdicts, confirmed, blocked, not found in scope, or insufficient evidence, and only a confirmed verdict counts as a violation. We then review every piece of evidence the agent flags. To prevent and detect Selection or Test instances obtained from the web or other sources, development blocks data sources that directly expose the task pools, and the audit scans downloads, training corpora, and code for protected task IDs, problem text, and n-gram fragments.

\begin{promptbox}[unbreakable]{Manual audit verdict template}
## Verdict (manual)

- cheating_confirmed: {cheating_confirmed}
- notes: {notes}
- decided_by: {decided_by}
\end{promptbox}

\paragraph{Findings.}
Of the 75 Qwen3.5-9B runs, 22 are confirmed violations. Twenty of them come from Kimi, Qwen, and GLM (7, 6, and 7 runs), Opus has 2, and Astra has none. All 22 runs accessed prohibited data, and 6 of them also trained on it (Table~\ref{tab:appendix-audit}). Static screening of the final harnesses finds no task-specific code or embedded test content in any run, so every violation entered only the training data, the weights, or the development process, and resetting the weights to Base removes the source of contamination.

\begin{table}[H]
\centering
\small
\begin{tabular}{lr}
\toprule
Check & Confirmed runs \\
\midrule
Accessing prohibited data & 22 \\
Training on prohibited data & 6 \\
Substituting unauthorized weights & 0 \\
Task-specific harness customization or embedded test content & 0 \\
\bottomrule
\end{tabular}
\caption{\textbf{Confirmed audit findings} over the 75 Qwen3.5-9B runs. Accessing prohibited data covers obtaining protected material, reading files outside the workspace, and accessing private evaluation code or external resources during evaluation.}
\label{tab:appendix-audit}
\end{table}

\paragraph{Strictest rule.}
Under the strictest rule, all 22 runs are reset to the initial system and keep no harness gains. The overall capability gap remains unchanged (Table~\ref{tab:appendix-strict}), and the gap between Opus and GLM widens from 12.0 to 13.7 points.

\begin{table}[H]
\centering
\small
\begin{tabular}{lrrrrr}
\toprule
Rule & Kimi & Qwen & GLM & Opus & Astra \\
\midrule
Main rule (Section~\ref{sec:evaluation-integrity}) & 7.93 & 9.40 & 12.24 & 24.19 & 31.23 \\
All 22 runs reset to the initial system & 5.45 & 7.60 & 8.29 & 21.94 & 31.23 \\
\bottomrule
\end{tabular}
\caption{\textbf{Weighted average Test performance (\%)} under the main and strictest rules.}
\label{tab:appendix-strict}
\end{table}